\documentclass[aos, preprint]{imsart}

\RequirePackage[OT1]{fontenc}
\RequirePackage{amsthm,amsmath}
\allowdisplaybreaks[4]

\usepackage{array}
\usepackage{mathrsfs}
\usepackage{amssymb,bbm}
\usepackage{amsfonts}
\usepackage{amsmath,amssymb}
\usepackage{graphicx}
\usepackage{amscd}
\usepackage{color}
\usepackage{mathrsfs}
\usepackage{latexsym,bm}
\usepackage{array}
\usepackage{threeparttable}
\usepackage{tikz}
\usetikzlibrary{arrows}
\usetikzlibrary{arrows.meta, calc}
\usetikzlibrary{shapes,positioning,decorations.markings}
\usetikzlibrary{arrows, automata, chains, positioning, shapes.geometric, shapes.symbols}
\usepackage{graphicx}
\usepackage{subcaption}
\usepackage{tikz-qtree}
\usepackage{tikz-qtree-compat}
\usepackage{color}
\usepackage{upgreek}
\usetikzlibrary{shapes,decorations,arrows,calc,arrows.meta,fit,positioning}
\tikzset{
	-Latex,auto,node distance =1 cm and 1 cm,semithick,
	state/.style ={ellipse, draw, minimum width = 0.7 cm},
	point/.style = {circle, draw, inner sep=0.04cm,fill,node contents={}},
	bidirected/.style={Latex-Latex,dashed},
	el/.style = {inner sep=2pt, align=left, sloped}
}

\usepackage[plain,noend]{algorithm2e}
\usepackage{extarrows}

\DeclareFontFamily{U}{mathx}{\hyphenchar\font45}
\DeclareFontShape{U}{mathx}{m}{n}{
      <5> <6> <7> <8> <9> <10>
      <10.95> <12> <14.4> <17.28> <20.74> <24.88>
      mathx10
      }{}
\DeclareSymbolFont{mathx}{U}{mathx}{m}{n}
\DeclareMathAccent{\widecheck}{\mathalpha}{mathx}{"71}

\startlocaldefs \numberwithin{equation}{section} \theoremstyle{it}
\newtheorem{thm}{Theorem}[section]
\newtheorem{lem}{Lemma}[section]

\endlocaldefs

\newtheorem{definition}{Definition}[section]
\newtheorem{example}{Example}[section]

\newcommand{\info}{\mathrm{\sigma}}

\newcommand{\indep}{\perp\mkern-9.5mu\perp}

\begin{document}

\begin{frontmatter}
\title{Info Intervention}
\runtitle{Info Intervention}

\begin{aug}
\author{\fnms{Heyang} \snm{Gong}\ead[label=e1]{gonghy@mail.ustc.edu.cn}}
\and
\author{\fnms{Ke} \snm{Zhu}\thanksref{t2}\ead[label=e2]{mazhuke@hku.hk}
}

\thankstext{t2}{Supported in part by RGC of Hong Kong (Nos. 17306818 and
17305619) and National Natural Science Foundation of China (Nos. 11571348, 11371354 and 71532013)}
\runauthor{Gong and Zhu}

\affiliation{University of
Science and Technology of China \& University of Hong Kong}

\address{University of
Science and Technology of China\\
Department of Mathematics\\
Hefei, Anhui\\
China\\
\printead{e1}}

\address{University of Hong Kong\\
Department of Statistics \\
\,\,\,\,\& Actuarial Science\\
Pok Fu Lam Road\\
Hong Kong\\
\printead{e2}}
\end{aug}

\begin{abstract}
Causal diagrams based on \emph{do} intervention are useful tools to formalize, process and understand
causal relationship among variables. However, the \emph{do} intervention has
controversial interpretation of causal questions for non-manipulable variables, and it also lacks the power to check the
conditions related to counterfactual variables. This paper introduces a new \emph{info} intervention to tackle these two problems, and
provides causal diagrams for communication and theoretical focus based on this \emph{info} intervention.
Our \emph{info} intervention intervenes the input/output information of causal mechanisms, while the \emph{do} intervention intervenes the causal mechanisms.
Consequently, the causality is viewed as information transfer in the \emph{info} intervention framework.
 As an extension, the generalized  \emph{info} intervention is also proposed and studied in this paper.
\end{abstract}


\begin{keyword}
\kwd{Causal calculus; Directed acyclic graph; \emph{do} intervention; \emph{info} intervention; Structural causal model} 
\end{keyword}

\end{frontmatter}

\section{Introduction}

Since the seminal work of Pearl (1995), the causal diagrams based on \emph{do} intervention have
been an important tool for causal inference. Pearl's causal diagrams
provide not only a formal language for communicating causal questions but also an effective way for
identifying causal effects. These merits of causal diagrams have stimulated
applications in many fields; see, e.g., Greenland, Pearl and Robins (1999), White and Lu (2011),
Meinshausen et al. (2016), Williams et al. (2018), and Hunermund and Bareinboim (2019) among others. In the era of big data and AI,
three fundamental obstacles are standing in our way to strong AI, including robustness (or adaptability),
explainability and lacking of understanding cause-effect connections (Pearl, 2019a),
and the hard open problems of machine learning are intrinsically related to causality (Sch\"{o}lkopf, 2019).
Encouragingly, Pearl (2019a) asserted that all these obstacles can be overcome using causal modeling tools, in particular, causal diagrams
and their associated logic.

Pearl's causal diagrams based on \emph{do} intervention are mainly manipulated in the causal directed acyclic graph (DAG),
under which Markov factorization is assumed for the joint distribution of all variables, giving rise a
way to calculate the intervention distribution
in \emph{do}-intervention DAG.
In a causal DAG,
Pearl's  diagrams can identify the causal effects by using observational probabilities
under several conditions (e.g., ``back-door''/``front-door'' criteria and Pearl's three rules), which
could largely demystify the haunting ``confounding'' problem in applications.
Although the \emph{do} intervention provides an effective way for causal inference,
 criticisms of this operator still exist. One is that the empirical interpretation of
 \emph{do} intervention is controversial when applied to non-manipulable variables such as age, race, obesity, or cholesterol level (Pearl, 2019b).
 Another one is that the counterfactual variables (also known as the potential outcome variables)  are not included on causal diagrams, and so the conditions related
  to counterfactual variables (e.g., the conditional independence and ``ignorability'' conditions) can not be directly read off the graph (Hern\'{a}n and Robins, 2019).

In this paper, we first introduce a new \emph{info} intervention for the structural causal model (SCM), which nests
the DAG as a special case. Let $X$ be the intervention variable in SCM.
Our \emph{info} intervention operator $\info(X=\tilde{x})$ sends out the
information $X=\tilde{x}$ to the descendant nodes of $X$, while keeping the rest of the model intact. Since
the information $X=\tilde{x}$ has been received by its descendant nodes, the edges as communication channels
from $X$ to its child nodes are removed after \emph{info} intervention.
In other words, the \emph{info} intervention replaces $X$ by $\tilde{x}$ in the structural equations of
child nodes of $X$, and then updates the input of structural equations at other
descendant nodes of $X$ accordingly.
Consequently, the variables at descendant nodes of $X$
become counterfactual variables, and they have no causal effect from $X$ in the \emph{info}-intervention SCM.
This is different from Pearl's \emph{do} intervention operator $do(X=\tilde{x})$,
under which the value of $X$ is forced to be a hypothetical value $\tilde{x}$, and the edges of $X$ from its parent nodes are removed,
while keeping the rest of the model intact.
Owing to this difference, the causal questions on non-manipulable variables are not controversial
for the \emph{info} intervention, because the \emph{info} intervention keeps the mechanisms of $X$, and changes the variables at descendant nodes of $X$ to  counterfactual variables with the transferred information
$X=\tilde{x}$; at the same time, since the counterfactual variables exist
in the \emph{info}-intervention SCM, some conditions on counterfactual variables could be directly read off the graph
in many circumstances.


Next, we present a new \emph{info}-causal DAG based on \emph{info} intervention.
Our \emph{info}-causal DAG has the same graph for all considered variables as Pearl's causal DAG, and
it also follows the causal DAG to assume Markov factorization for
the joint distribution of all variables. However, unlike the causal DAG,
the \emph{info}-causal DAG specifies the intervention distribution differently.
The first difference is that the intervention variables
inheriting their pre-intervention distributions are still random  in \emph{info}-intervention DAG,
while they become deterministic
in \emph{do}-intervention DAG. This difference is compatible to the fact that
the \emph{info} intervention keeps all causal mechanisms as in ``Law-like'' causality framework,
and the \emph{do} intervention removes the causal mechanisms of intervention variables.
The second difference is that the intervention distribution in \emph{info}-intervention DAG is for
a group of variables including either those from the \emph{info}-causal DAG or
the counterfactual ones (i.e., the variables at descendant nodes of intervention variables), however,
the intervention distribution in \emph{do}-intervention DAG is always for all variables
from the causal DAG. This difference makes the \emph{info} intervention being capable to deal with
counterfactual variables as in experimental causality framework.

Due to the aforementioned two differences, our \emph{info}-causal DAG could have several advantages over
Pearl's causal DAG. First, the \emph{info}-causal DAG  inherits the advantage of \emph{info} intervention SCM
to avoid the controversial interpretation of the questions for non-manipulable variables
and check the conditions for counterfactual variables.
Second, the \emph{info}-causal DAG can raise
the interventional and counterfactual questions conditional on the intervention variables, while the causal DAG can not.
This advantage is particularly important when the intervention variables still have the causal effects on
its (part of) descendants after intervention. In this case, we can naturally extend our  \emph{info} intervention idea
to form a so-called generalized \emph{info}-causal DAG, however, this seems challenging in the \emph{do} intervention framework.


Interestingly, compared with Pearl's causal DAG, our \emph{info}-causal DAG not only exhibits the advantage in terms of
communication, but also keeps the capability in terms of theoretical focus. Specifically, we show that the
causal calculus under some conditions (e.g., ``back-door''/``front-door'' criteria and three rules)
shares the same formulations expressed by observational probabilities in both DAGs.
Although the causal calculus has no difference in form, the underlying causality is essentially different in
both DAGs. In the \emph{info}-causal DAG, we view the causality as information transform, meaning that
a variable $X$ causes another variable $Y$ iff ``changing information on $X$ leads to a potential change in $Y$, while keeping everything else constant''.
In the causal DAG, the causality is viewed differently, saying that  $X$ causes $Y$ iff ``changing $X$ leads to a change in $Y$, while keeping everything else constant''.


We shall mention that our \emph{info}-causal DAG has certain similarities with the
single-world intervention graph (SWIG) in Richardson and Robins (2013).
The SWIG is an approach to unifying graphs and counterfactuals via splitting
every intervention node into a random node and a fixed node.
Although both SWIG and \emph{info}-causal DAG consider the counterfactual variables, they
have several differences. First, the \emph{info}-intervention DAG does not contain fixed nodes,
leading to a more neat way for use than the SWIG.
Second,
the intervention distribution in \emph{info}-intervention DAG is determined by the joint distribution
in \emph{info}-causal DAG, while that in SWIG is directly assumed. Consequently,
in terms of causal calculus, the SWIG needs additional modularity assumptions to link the conditional distribution of
counterfactual variables in intervention DAG to that of corresponding
 variables in observational DAG.
 Third, the systematic tools in terms of communication and theoretical focus
 as Pearl's causal DAG are present for the \emph{info}-causal DAG but absent for the SWIG.
Fourth, the \emph{info}-causal DAG is extended to the generalized \emph{info}-causal DAG,
and this extension seems hard for the SWIG.



The remainder of this paper is organized as follows. Section 2 gives the preliminaries on Pearl's diagrams.
Section 3 introduces our \emph{info} intervention. Section 4 presents our \emph{info}-causal DAG.
Section 5 provides the causal calculus based on the \emph{info}-causal DAG. An extension work on the generalized \emph{info} intervention is considered in Section 6.
Concluding remarks are offered in Section 7. Proofs
are relegated to the Appendix.

\section{Preliminaries}

The notation of causality has been much examined, discussed and debated in science and philosophy over many centuries. Among many proposed frameworks for causality, Pearl's causal diagrams have been widely used in the real world to formalize causal questions and implement causal inferences for observational data. In this section, we mainly introduce some preliminaries on Pearl's causal diagrams, which are build on the structural causal model (SCM) (also known as structural equation model) to make graphical assumptions of the underlying data generating process (see, e.g., Pearl (2009) and Forr\'{e} and Mooij (2019)  for overviews).

\begin{definition}[SCM]
	\label{SCM-def}
	An SCM by definition consists of:
	\begin{enumerate}
		\setlength\itemsep{0em}
		\item A set of nodes $V^+=V \dot \cup U$, where elements of $V$ correspond to endogenous variables, elements of $U$ correspond to exogenous (or latent) variables, and
		$V \dot \cup U$ is the disjoint union of sets $V$ and $U$.
		
		\item An endogenous/exogenous space $\mathcal{X}_v$ for every $v \in V^+$, $\mathcal{X}:=\prod_{v \in V^+}\mathcal{X}_v$.
		
		\item A product probability measure $P:=P_U=\otimes_{u \in U} P_u$ on the latent space $\prod_{u \in U} \mathcal{X}_u$.
		
		\item A directed graph structure $G^+=(V^+, E^+)$, with a set of directed edges $E^+$ and a system of structural equations  $f_V = (f_v)_{v \in V}$:
		\[f_v: \; \prod_{s \in  pa(v)\cap G^+}\mathcal{X}_s \to \mathcal{X}_v,\]
		where $ch(U)\cap G^+ \subseteq V$, all functions $f_V$ are measurable, and $ch(v)$ and $pa(v)$ stand for child and parent nodes of $v$ in $G^+$, respectively.
	\end{enumerate}
	Conventionally, an SCM can be summarized by the tuple $M=(G^+,\mathcal{X},P, f)$. Note that $G^+$ is referred as the augmented functional graph, while the functional graph which includes only endogenous variables, is denoted as $G$.
\end{definition}

According to its definition, the SCM deploys three parts, including graphical models, structural equations, and counterfactual and interventional logic. Graphical models serve as a language for representing what we know about the world, counterfactuals help us to articulate what we want to know, while structural equations serve to tie the two together in solid semantics.

Let $X_{A}$ be a set of variables at the nodes $A$. For any $I \subseteq V$,
the key implementation of Pearl's causal diagrams is to capture interventions by using an intervention operator called  $do(X_{I}=\tilde{x}_{I})$, which
simulates physical interventions by deleting certain functions from the model, replacing them with a constant vector $X_{I}=\tilde{x}_{I}$, while keeping the rest of the model unchanged.
Formally, $X_I$ are called intervention variables, and this \emph{do} intervention operator on $X_{I}$ is defined as follows:

\begin{definition}[\emph{do} intervention]
	\label{def:p-intervention}
	Given an SCM $M=(G^+,\mathcal{X},P, f)$ for $X_{V}$ and any $I \subseteq V$, the do intervention $do(X_I=\tilde{x}_I)$ (or, in short, $do(\tilde{x}_I)$) maps $M$ to the do-intervention model $M^{do(\tilde{x}_I)} = (G^+, \mathcal{X}, P, \tilde{f})$ for $X_V$, where
	$$
	\begin{aligned}
    \tilde f_v(X_{pa(v)\cap V}, X_{pa(v)\cap U}) := \begin{cases}
	\tilde{x}_v, & v \in I, \\
	f_v(X_{pa(v)\cap V}, X_{pa(v)\cap U}), & v \in V \setminus I \,.
	\end{cases}
	\end{aligned}
	$$
%
\end{definition}


Since there has theoretical and technical complications in dealing with the cyclic SCM, most of efforts are made to study the directed acyclic graph (DAG) in the class of
acyclic SCMs (Bongers et al., 2020).
Generally speaking, the DAG can be viewed as the non-parametric analogue of an
acyclic SCM. Denote a DAG by $G=(V, E)$, with a set of nodes $V$ and a set of directed edges $E$.
For ease of notation, we re-define $X:=X_{V}$ as the variables at $V$. To do causal inference in $G$,
we need specify a way to calculate the intervention distribution of $X$ in the \emph{do}-intervention DAG, and this leads to
the so-called causal DAG, under which the causal semantics could be well defined without any complications.



\begin{definition}[Causal DAG]
	Consider a DAG $G=(V, E)$ and a random vector $X$ with distribution $P$.
Then, $G$ is called a causal DAG for $X$ if $P$ satisfies the following:
	\begin{enumerate} 
		\setlength\itemsep{0em}
		\item $P$ factorizes, and thus is Markov, according to $G$, and
		\item for any $A \subseteq V$, $B = V/A$,  and any $\tilde{x}_A, x_B$ in the domains of $X_A, X_B$,
		\begin{equation}\label{eq:sigma:do}
P(x	| do(\tilde{x}_A)) = \prod_{k \in B} P(x_k|x_{pa(k)})  \prod_{j \in A} \mathbb{I}(x_j = \tilde{x}_j).
		\end{equation}
	\end{enumerate}
\end{definition}

In view of (\ref{eq:sigma:do}), the difference between the observational distribution
$P(x)$ and the \emph{do}-intervention distribution $P(x| do(\tilde{x}_A))$ is that all factors $P(x_j |x_{pa(j)})$, $j\in A$, are
removed and replaced by degenerate probabilities $\mathbb{I}(x_j = \tilde{x}_j)$, while all remaining factors
$P(x_k |x_{pa(k)})$, $k\in B$, stay the same.

For a causal DAG $G$, its \emph{do}-intervention DAG is defined as follows:

\begin{definition}[\emph{do}-intervention DAG]
	Consider a causal DAG $G = (V, E)$ for a random vector $X$, and its do intervention $do(\tilde{x}_A)$. Then, the do-intervention DAG, denoted by $G^{do(\tilde{x}_A)}$, is for $X$, which has the do-intervention distribution $P(x| do(\tilde{x}_A))$ in (\ref{eq:sigma:do}).
\end{definition}

In a causal DAG, if
$P(x_{B}| do(\tilde{x}_A))\not=P(x_{B}| do(\tilde{x}_A'))$ for two values $\tilde{x}_{A}\not=\tilde{x}_{A}'$,
we say that $X_{A}$ has the causal effect on $X_{B}$.
Clearly, to draw the causal inference, it is crucial to determine whether $P(x_{B}| do(\tilde{x}_A))$
can be calculated by using the observational probabilities.
Pearl (1995) raised the concept of ``identifiability'' to answer this question.

\begin{definition}[Identifiability]
	The causal effect of $X_{A}$ to $X_{B}$ is said to be identifiable if the quantity $P(x_{B}| do(\tilde{x}_A))$
can be computed uniquely from any positive distribution of the observed variables that is compatible with $G$.\label{identi_def}
\end{definition}

To check the identifiability, Pearl (1995) gave ``back-door''/``front-door'' criteria and three rules, which are widely used to deal with the ``confounding'' problem, making the causal DAG attractive in plenty of applications. Once the causal effect is identifiable,
many quantities of interest (e.g., the average causal effect (ACE) and the conditional ACE) can be calculated accordingly.

Besides Pearl's causal diagrams, many other DAGs or graphical models can also be used to implement the causal inference. For more discussions and developments in this context, we refer to Lauritzen and Richardson (2002), Richardson and Spirtes (2002), Peters et al. (2016), Maathuis et al. (2018), Rothenhausler, B\"{u}hlmann and Meinshausen (2019), and references therein.

%

\section{Info intervention}
\label{sec:info}

Causal questions, such as what if we make something happen, can be formalized by using $do$ intervention, which, however, is still controversial on empirical understanding as
mentioned by Pearl (2018, 2019b).
In many settings, a \emph{do} intervention which forces the variable to a given value is somewhat idealized or hypothetical.
For instance, it is controversial to manipulate variables such as age, race, obesity, or cholesterol level by setting their values to some hypothetical ones in \emph{do} intervention.
To see this point more clearly, we consider the following example.
\begin{example} 
    \label{eg:real}
	Assume a causal relationship among 6 domain variables in Fig\,\ref{fig:exercise}.
	Suppose that we aim to study the causal effect of Age on Income by using do intervention.
	However, the empirical interpretation of the intervention $do(Age = \tilde{a})$
	is controversial for a hypothetical positive integer $\tilde{a}$, though Pearl suggested that we should interpret it in other dimensions (Pearl, 2019b). In other words, it seems unreasonable to articulate a causal question that what if Age equals $\tilde{a}$ under do intervention framework, indicating that the do intervention is not complete for formalizing real world causal questions.
    \begin{figure}[ht]
        \label{fig:real}
    	\centering
   	\begin{tikzpicture}
    	\node[state] (O) at (0.0, 2.0) {$Occupation$};
    	\node[state] (d) at (0,0) {$Exercise$};
     	\node[state] (I) at (2.9, 2.0) {$Income$};
    	\node[state] (x) at (2.3,0.8) {$Age$};
    	\node[state] (D) at (4.5, 1.1) {$Diet$};
    	\node[state] (y) at (4.5,0) {$Cholesterol$};
    	\path (x) edge (d);
    	\path (x) edge (y);
    	\path (d) edge (y);
    	\path (O) edge (d);
    	\path (x) edge (I);
    	\path (O) edge (I);
    	\path (I) edge (D);
    	\path (D) edge (y);
    	\end{tikzpicture}
    	\caption{A causal relationship}
    	\label{fig:exercise}
    \end{figure}
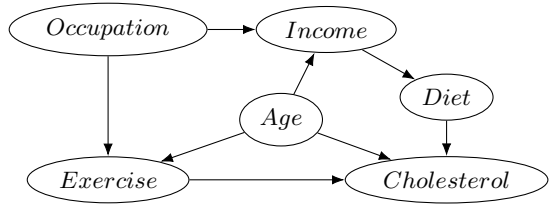
\end{example}

Although it is controversial to  force or set \emph{Age} to a hypothetical  value $\tilde{a}$ in Example \ref{eg:real}, it is always non-controversial  to send out the information that \emph{Age} is $\tilde{a}$. This motivates us to consider new causal semantics based on \emph{info} intervention operator, whose formal definition is given below.

\begin{definition}[\emph{Info} intervention]
	\label{def:i-intervention}
	Given an SCM $M=(G^+,\mathcal{X},P, f)$ for $X_V$ and any $I \subseteq V$, the info intervention $\info(X_I=\tilde{x}_I)$ (or, in short, $\info(\tilde{x}_I)$) maps $M$ to the info-intervention model $M^{\sigma(\tilde{x}_I)} = (G^+, \mathcal{X}, P, f)$ for
$X^{\info(\tilde{x}_I)}_V$, where
    $$
    X^{\info(\tilde{x}_I)}_v= {f}_v(\widetilde{X}_{V \cap pa(v)}, X_{U \cap pa(v)})
    $$
   with $\widetilde{X}_j = \tilde{x}_j$ if $j\in I$ else $X^{\info(\tilde{x}_I)}_j$.
\end{definition}

Let $desc(I)$ denote the descendant nodes of every node in $I$.
Based on Definition \ref{def:i-intervention}, we can show that for any node $i\not\in desc(A)$ with $A\subseteq V$, $X_i^{\info(\tilde{x}_{A})}=X_i$. Also,
for two disjoint sets $A, B\subseteq V$,
$X^{\info(\tilde{x}_{A}, \tilde{x}_{B})}_v:=\big(X^{\info(\tilde{x}_{A})}_v\big)^{\info(\tilde{x}_{B})}$ has the
commutative property, that is, $X^{\info(\tilde{x}_{A}, \tilde{x}_{B})}_v=X^{\info(\tilde{x}_{B}, \tilde{x}_{A})}_v$ for all $v\in V$.

Moreover, based on Definition \ref{def:i-intervention}, we know that	
the \emph{info}-intervention SCM $M^{\info(\tilde{x}_I)}$
does not delete any structural equations $f_{V}$ from the model, but just sends out the information
$X_I=\tilde{x}_I$ to $desc(I)$. Since the information $X_I=\tilde{x}_I$ has been received by
$desc(I)$, the edges from $I$ to $ch(I)$ (i.e., the child nodes of $I$) are removed in $M^{\info(\tilde{x}_I)}$, and
the variables at $desc(I)$ become counterfactual variables
with a hypothetical input $X_I=\tilde{x}_I$ in their structural equations.
Note that the counterfactual variables are also known as the potential outcome variables, and they
are unobservable variables that live in the counterfactual world.
For more discussions on the potential outcome framework, we refer to Rubin (1974), Angrist, Imbens and Rubin (1996),
Imbens and Rubin (2015), Imbens (2019) and references therein.

Compared with the \emph{do} intervention $do(\tilde{x}_I)$, the \emph{info} intervention $\info(\tilde{x}_I)$ has two critical differences.
First, $M^{\info(\tilde{x}_I)}$ keeps the causal mechanisms (i.e., the structural equations $f_{V}$) unchanged, while $M^{do(\tilde{x}_I)}$ does not.
Second, $M^{\info(\tilde{x}_I)}$ contains the counterfactual variables at $desc(I)$, while $M^{do(\tilde{x}_I)}$ does not. The first difference makes the \emph{info} intervention have no non-manipulable variables problem, which, however, exists
in the \emph{do} intervention. For example, we can articulate
the causal question that what if \emph{Age} equals $\tilde{a}$ in Example \ref{eg:real} by using $\info(\tilde{a}):=\info(Age=\tilde{a})$, since
the descendant nodes of \emph{Age} in $M^{\info(\tilde{a})}$ do receive the hypothetical value \emph{Age}\,$=\tilde{a}$, and at the same time,  the actual value of this intervention variable \emph{Age} in $M^{\info(\tilde{a})}$ does not change.
The second difference could provide us a direct visual way to determine the conditional independence
between variables in $M$ and counterfactual variables in $M^{\info(\tilde{x}_I)}$, however, this is hardly feasible
in the \emph{do} intervention framework. Re-consider Example \ref{eg:real}, in which we can easily use the $d$-separation argument to show that given \emph{Occupation}, \emph{Exercise}$^{\info(\tilde{a})}$ and \emph{Income}$^{\info(\tilde{a})}$ are independent. Clearly, this relationship can not be found by using the \emph{do} intervention.

To further illustrate how the \emph{info} intervention works and what are the differences between \emph{info} and \emph{do} interventions, we consider the following example:

\begin{figure}[htp]
	\begin{subfigure}[tb]{1.0\textwidth}
		\centering
		\begin{tikzpicture}
		\node[state] (d) at (0,0) {$T$};
		\node[state] (ed) at (-0.5, 1) {$\epsilon_T$};
		\node[state] (x) at (0.5, 1.5) {$Z$};
		\node[state] (ex) at (2.2,1.5) {$\epsilon_Z$};
		\node[state] (y) at (2,0) {$Y$};
		\path (ex) edge (x);
		\path (x) edge (d);
		\path (ex) edge (x);
		\path (ed) edge (d);
		\path (d) edge (y);
		\path (x) edge (y);
		\end{tikzpicture}
		\caption{An SCM $M$.}
		\label{fig:SCM1}
	\end{subfigure}
	\vfill
	\begin{subfigure}[tb]{0.4\textwidth}
		\centering
		\begin{tikzpicture}
		\node[state] (d) at (0,0) {$T=\tilde{t}$};
		\node[state] (x) at (0.5,1.5) {$Z$};
		\node[state] (y) at (2, 0) {$Y$};
		\node[state] (ex) at (2.0,1.5) {$\epsilon_Z$};
		\node[state] (ed) at (-0.5, 1) {$\epsilon_T$};
		\path (ex) edge (x);
		\path (x) edge (y);
		\path (d) edge (y);
		\end{tikzpicture}
		\caption{$M^{\emph{do}(\tilde{t})}$.}
		\label{fig:SCM2}
	\end{subfigure}
	\begin{subfigure}[tb]{0.4\textwidth}
		\centering
		\begin{tikzpicture}
		\node[state] (d) at (0,0) {$T$};
		\node[state] (x) at (0.5,1.5) {$Z$};
		\node[state] (y) at (2, 0) {$Y^{\info(\tilde{t})}$};
		\node[state] (ex) at (2.0,1.5) {$\epsilon_Z$};
		\node[state] (ed) at (-0.5, 1) {$\epsilon_T$};
		\path (ex) edge (x);
		\path (ed) edge (d);	
		\path (x) edge (y);
		\path (x) edge (d);
		\end{tikzpicture}
		\caption{$M^{\sigma(\tilde{t})}$.}
		\label{fig:SCM3}
	\end{subfigure}
	\caption{An SCM and its two intervention SCMs.}
	\label{fig:info-VS-$do$}
\end{figure}
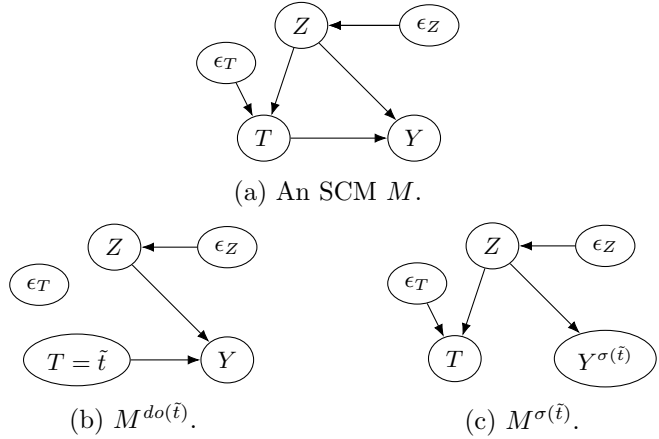

\begin{example}
	\label{eg1}
	An SCM $M$ with
	a treatment $T$, an outcome $Y$, a confounder $Z$,
	and two latent variables $\epsilon_T, \epsilon_Z$ is given in Fig\,\ref{fig:info-VS-$do$}(a), and its structural equations are:
	$$
	\begin{aligned}
	\begin{cases}
	Z = f_Z(\epsilon_Z),  \\   
	T = f_T(Z, \epsilon_T), \\
	Y = f_Y(T, Z).
	\end{cases}
	\end{aligned}
	$$
	Based on Definition \ref{def:p-intervention}, its
	do-intervention SCM $M^{do(\tilde{t})}$ given in Fig\,\ref{fig:info-VS-$do$}(b) has the following structural equations:
	$$
	\begin{aligned}
	\begin{cases}
	Z = f_Z(\epsilon_Z),  \\
	T = \tilde{t}, \\
	Y = f_Y(T, Z).
	\end{cases}
	\end{aligned}
	$$    	
	Based on Definition \ref{def:i-intervention}, its info-intervention SCM $M^{\info(\tilde{t})}$ given in Fig\,\ref{fig:info-VS-$do$}(c) has the following structural equations:
	$$
	\begin{aligned}
	\begin{cases}
	Z = f_Z(\epsilon_Z),  \\
	T = f_T(Z, \epsilon_T), \\
	Y^{\info(\tilde{t})} = f_Y(\tilde{t}, Z),
	\end{cases}
	\end{aligned}
	$$
where we have used the fact that $Z^{\info(\tilde{t})}=Z$ and $T^{\info(\tilde{t})}=T$.
Note that the causal mechanisms (i.e., $f_{Z}$, $f_{T}$ and $f_{Y}$)
are unchanged only in $M^{\info(\tilde{t})}$, while $Y$ in $M$ becomes a counterfactual variable $Y^{\info(\tilde{t})}$ in $M^{\info(\tilde{t})}$.
Moreover, from Fig\,\ref{fig:info-VS-$do$}(c), the $d$-seperation argument (Geiger, Verma and Pearl, 1990) implies that given $Z$, $Y^{\info(\tilde{t})}$ and $T$ are independent. That is, the ignorability condition of Rosenbaum and Rubin (1983) can be directly read off Fig\,\ref{fig:info-VS-$do$}(c).
\end{example}

In the \emph{do} intervention framework, Pearl (2019a) raised the concept of \emph{three-level causal hierarchy} to articulate the causal questions into three levels: 1. Association; 2. Intervention; 3. Counterfactuals.
This classification of causal questions gives us a useful insight on what kind of questions each class is capable of answering, and questions at level $i$ can be answered only if information from level $j$ $(>i)$ is available.
Similar to Pearl's idea, we can give a three-level causal hierarchy in the \emph{info} intervention framework:
\begin{enumerate} 
	\setlength{\itemsep}{-1pt}
	\item Association $P(x_B|x_A):=P(X_B=x_B|X_A=x_A)$
	\begin{itemize}
		\setlength{\itemsep}{0pt}
        \item Typical activity: Seeing.
		\item Typical questions: What is? How would seeing $X_A$ change my belief in $X_B$?
		\item Examples: What does the habit of exercise information tell me about the cholesterol level?
              What does a symptom tell me about a disease?
	\end{itemize}
	\item Intervention $P(x_B|\sigma(\tilde{x}_A), x_{C}):=P(X_B^{\sigma(\tilde{x}_A)}=x_B|X_C^{\sigma(\tilde{x}_A)}=x_C)$
	\begin{itemize}
		\setlength{\itemsep}{0pt}		
		\item Typical activity: Intervening.
		\item Typical questions: What if?  What if I manipulate the information sending out from $X_A$?
		\item Examples: What will the income be if the company received the information that my age is 32?
              What will happen if the public received the information that the price is doubled?
	\end{itemize}
	\item Counterfactuals $P(x_B^{\sigma(\tilde{x}_A)}|x_A, x_B):=P(X_B^{\sigma(\tilde{x}_A)}=x_B^{\sigma(\tilde{x}_A)}|X_A=x_A, X_B$ $=x_B)$
	\begin{itemize}
		\setlength{\itemsep}{0pt}		
		\item Typical activity: Imagining, Retrospection.
		\item Typical questions: Why?  Was it the information of $X_A$ that caused $X_B$?
		\item Examples: Was it the information of young age caused me to have low income?  What if I had told the company the information my age is 32 given that my actual age is 22?
	\end{itemize}
\end{enumerate}

Both Pearl's and our three-level causal hierarchies can formulate kinds of causal questions.
One advantage of our three-level causal hierarchy is that the causal questions for non-manipulable variables are not controversial.
This is due to the fact that the \emph{info} intervention does not change the causal mechanisms of the intervention variables, but just sends out the information on the intervention to the model.
To make this point more clearly, we re-consider Example \ref{eg:real}. In this example,
the interventional question, What will be my income if I force my age to 32,
i.e., $P(\mathit{Income}|do(\mathit{Age}=32))$, is controversial. In contrast, the interventional question
$P(\mathit{Income}| \sigma(\mathit{Age}=32))$ can be interpreted as what will be my income if the company receives
the information that my age is 32, and the counterfactual question $P(\mathit{Income}^{\sigma(\mathit{Age}=32)} | \mathit{Age}=22, \mathit{Income}=low)$
can be interpreted as what would happen to my income if the company had received the information my age is 32 given that my
actual age is 22 and my actual income is low.

\section{Info-causal DAG}

For a given set of variables $X$, the causal DAG is a useful tool to study their causal relationship in the framework of
\emph{do} intervention. Analogous to the causal DAG, it is natural to study the causal relationship in $X$ by introducing our \emph{info}-causal DAG below, under which the causal semantics could be well defined without any complications in the framework of \emph{info} intervention.

\begin{definition}[\emph{Info}-causal DAG]
	Consider a DAG $G = (V, E)$ and a random vector $X$ with distribution $P$.
Then, $G$ is called an info-causal DAG for $X$ if $P$ satisfies the following:
	\begin{enumerate} 
		\setlength{\itemsep}{0pt}
		\item $P$ factorizes, and thus is Markov, according to $G$,
		\item for any $A \subseteq V$ and any $\tilde{x}_A$ in the domains of $X_A$,
		\begin{equation}\label{eq:sigma}
			P(x| \sigma(\tilde{x}_A)) = \prod_{k \in V} P(x_k|x_{pa(k)}^*),
		\end{equation}		
		where $x^*_k = x_k$ if $k \notin A$ else $\tilde{x}_k$.
	\end{enumerate}\label{def_4_1}
\end{definition}

In view of (\ref{eq:sigma}), the difference between the observational distribution
$P(x)$ and the \emph{info}-intervention distribution $P(x| \sigma(\tilde{x}_A))$ is that
the factors $P(x_k |x_{pa(k)})$ satisfying $pa(k)\cap A\not=\emptyset$ in $P(x)$, are replaced by
$P(x_k |x_{pa(k)}^{*})$ in $P(x| \sigma(\tilde{x}_A))$ with
$x_j$, $j\in pa(k)\cap A$, replaced by $\tilde{x}_j$, while
all remaining factors $P(x_k |x_{pa(k)})$ satisfying $pa(k)\cap A=\emptyset$ in $P(x)$,
are unchanged after \emph{info} intervention.

\subsection{Info-intervention DAG}

For a causal DAG, its \emph{do}-intervention DAG could provide possible simple graphical tests to identify the causal (Pearl, 1995).
For our \emph{info}-causal DAG $G$, its \emph{info}-intervention DAG can not only
work as the \emph{do}-intervention DAG, but also provide a visible way to check the conditions
for counterfactual variables. Our \emph{info}-intervention DAG is formally defined as follows:

\begin{definition}[\emph{Info}-intervention DAG]
	Consider an info-causal DAG $G = (V, E)$ for a random vector $X$, and its info intervention $\sigma(\tilde{x}_A)$.
The info-intervention DAG, denoted by $G^{\sigma(\tilde{x}_A)}$, is for $X^{\sigma(\tilde{x}_{A})}$, which has the info-intervention distribution $P(x| \sigma(\tilde{x}_A))$ in (\ref{eq:sigma}), where $X^{\sigma(\tilde{x}_{A})}$ is defined in the same way as $X$, except that the variables at descendant nodes of $A$ (say, $X_{desc(A)}$) are replaced
by the counterfactual variables (say, $X_{desc(A)}^{\sigma(\tilde{x}_{A})}$).\label{info_inter_dag}
\end{definition}

Due to the distinct forms of intervention distribution, our \emph{info}-intervention DAG $G^{\sigma(\tilde{x}_A)}$ has
three differences from Pearl's \emph{do}-intervention DAG $G^{do(\tilde{x}_A)}$. First,
$G^{\sigma(\tilde{x}_A)}$ is for $X^{\sigma(\tilde{x}_{A})}$, which is
a union of $X_{V/des(A)}$ and $X_{desc(A)}^{\sigma(\tilde{x}_{A})}$. Here, $X_{V/desc(A)}$ are part of $X$, and
$X_{desc(A)}^{\sigma(\tilde{x}_{A})}$ are counterfactual variables carrying the information $X_A=\tilde{x}_{A}$.
On the contrary, $G^{do(\tilde{x}_A)}$ is still for $X$.
This difference makes \emph{info}-intervention $\sigma(\tilde{x}_A)$ answer the causal questions of
non-manipulable variables non-controversially and check the conditions related to counterfactual variables visibly,
as demonstrated in Example \ref{eg:real} above.

Second, the \emph{info}-intervention distribution $P(x| \sigma(\tilde{x}_A))$
is different from the \emph{do}-intervention distribution $P(x| do(\tilde{x}_A))$. The factors  $P(x_k |x_{pa(k)})$, $k\in A$, in
$P(x| \sigma(\tilde{x}_A))$ inherit from those in $P(x)$, however, they are set to be
degenerate probabilities in  $P(x| do(\tilde{x}_A))$. This means that the intervention variables $X_A$ are forced to
hypothetical values $\tilde{x}_A$ in $G^{do(\tilde{x}_A)}$, while their joint distribution in $G^{\sigma(\tilde{x}_A)}$ has
the same form as that in $G$. This difference is compatible to that exhibited in SCM, where the \emph{info} intervention keeps
the causal mechanisms of endogenous variables, but the \emph{do} intervention forces the intervention endogenous variables to
hypothetical values (see Example \ref{eg1} above for an illustration).

\begin{figure}[htp]
	\begin{subfigure}[tb]{1.0\textwidth}
		\centering
		\begin{tikzpicture}
		\node[state] (d) at (0,0) {$X_A$};
		\node[state] (x) at (1, 1.5) {$X_D$};
		\node[state] (y) at (2, 0) {$X_B$};
		\node[state] (c) at (3.5, 0) {$X_C$};
		\path (x) edge (d);
		\path (d) edge (y);
		\path (x) edge (y);
		\path (y) edge (c);
		\end{tikzpicture}
		\caption{A DAG $G$.}
		\label{fig:dag1}
	\end{subfigure}
	\vfill
	\begin{subfigure}[tb]{0.42\textwidth}
		\centering
		\begin{tikzpicture}
		\node[state] (d) at (0,0) {$X_A=\tilde{x}_A$};
		\node[state] (x) at (1, 1.5) {$X_D$};
		\node[state] (y) at (2,0) {$X_B$};
		\node[state] (c) at (3.5, 0) {$X_C$};
		\path (d) edge (y);
		\path (x) edge (y);
		\path (y) edge (c);
		\end{tikzpicture}
		\caption{$G^{\emph{do}(\tilde{x}_A)}$.}
		\label{fig:dag2}
	\end{subfigure}
	\begin{subfigure}[tb]{0.42\textwidth}
		\centering
		\begin{tikzpicture}
		\node[state] (d) at (0,0) {$X_A^{\sigma(\tilde{x}_A)}$};
		\node[state] (x) at (1, 1.5) {$X_D$};
		\node[state] (y) at (2,0) {$X_B^{\sigma(\tilde{x}_A)}$};
		\node[state] (c) at (4.2, 0) {$X_C^{\sigma(\tilde{x}_A)}$};
		\path (x) edge (d);
		\path (x) edge (y);
		\path (y) edge (c);
		\end{tikzpicture}
		\caption{$G^{\sigma(\tilde{x}_A)}$.}
		\label{fig:dag3}
	\end{subfigure}
	\caption{A DAG and its two intervention DAGs.}
	\label{fig:info-do-dag}
\end{figure}
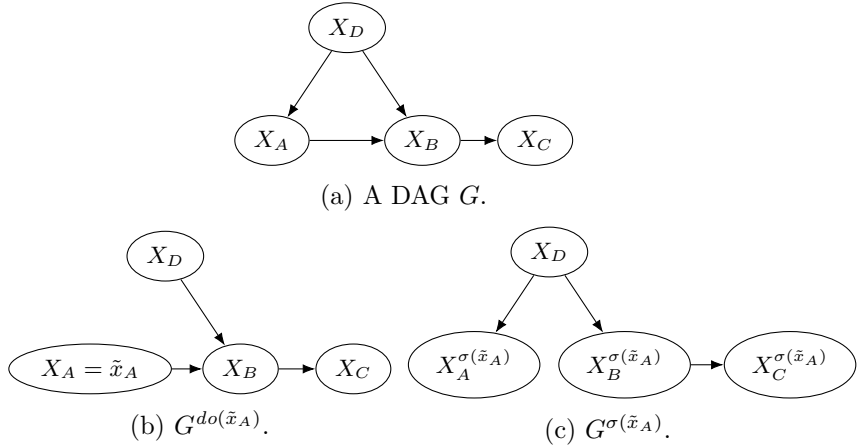

Third, $G^{\sigma(\tilde{x}_A)}$ and $G^{do(\tilde{x}_A)}$ are graphically different caused by
the difference between  $P(x| \sigma(\tilde{x}_A))$ and $P(x| do(\tilde{x}_A))$.
For simplicity, we assume no node in $A$ is in $desc(A)$.
In this case, the arrows from $pa(A)$ to $A$ are not removed by $\sigma(\tilde{x}_A)$,
and this ensures that the probability function of $X_A$ in $G^{\sigma(\tilde{x}_A)}$ has the same form as
that in $G$; however, these arrows are removed by $do(\tilde{x}_A)$, since $do(\tilde{x}_A)$ need
force $X_A$ to hypothetical values $\tilde{x}_A$. Meanwhile, the arrows from $A$ to $ch(A)$ are deleted by $\sigma(\tilde{x}_A)$, because the information $X_A=\tilde{x}_A$ has been received by $X_{ch(A)}^{\sigma(\tilde{x}_{A})}$;
however, these arrows are kept by $do(\tilde{x}_A)$ to capture how
$X_A$ affect other variables at $desc(A)$ in $G^{do(\tilde{x}_A)}$. To further illustrate this graphical difference, we
consider the following example:

\begin{example}
	\label{eg_dag}
	A DAG $G$ with four disjoint sets of variables $X_A$, $X_B$, $X_C$ and $X_D$ is given in
    Fig\,\ref{fig:info-do-dag}(a). Take $X_A$ as the intervention variables.
    Then, the do-intervention DAG $G^{do(\tilde{x}_A)}$ (see Fig\,\ref{fig:info-do-dag}(b)) removes the arrows from
    $pa(A)$ to $A$, and forces the intervention variables $X_A$ to take the hypothetical values $\tilde{x}_A$.
    On the contrary, the info-intervention DAG $G^{\info(\tilde{x}_A)}$ (see Fig\,\ref{fig:info-do-dag}(c))
    removes the arrows from $A$ to $ch(A)$, and forces each variable $X_i$ to be
    $X_{i}^{\sigma(\tilde{x}_{A})}$,
    where the variable $X_{i}^{\sigma(\tilde{x}_{A})}$ is a counterfactual variable if $i\in desc(A)$.
    In other words, $X_{D}^{\sigma(\tilde{x}_{A})}=X_D$ are not counterfactual variables,
    $X_{B}^{\sigma(\tilde{x}_{A})}$ and $X_{C}^{\sigma(\tilde{x}_{A})}$ are always counterfactual variables,
    and the variable $X_{i}^{\sigma(\tilde{x}_{A})}$, $i\in A$, is a counterfactual variable if $i\in desc(A)$.
\end{example}

\subsection{Causality as information transfer}

Consider two values $\tilde{x}_{A}\not=\tilde{x}_{A}'$. In the causal DAG,
$X_{A}$ has the causal effect on $X_{B}$, if $P(x_{B}| do(\tilde{x}_A))\not=P(x_{B}| do(\tilde{x}_A'))$.
In the \emph{info}-causal DAG, we say that $X_{A}$ has the causal effect on $X_{B}$, if
$P(x_{B}| \sigma(\tilde{x}_A))\not=P(x_{B}| \sigma(\tilde{x}_A'))$.
Moreover, we follow Definition \ref{identi_def} to say that
the causal effect of $X_A$ on $X_B$ is identifiable
if the quantity $P(x_{B}| \sigma(\tilde{x}_A))$
can be computed uniquely from any positive distribution of the observed variables that is compatible with $G$;
in this case, many quantities of interest (e.g., the ACE
and the conditional ACE) can be calculated accordingly.

It is worth noting that our viewpoint on causality is essentially different
from Pearl's, although the difference in form is just replacing Pearl's \emph{do} operator by our \emph{info} operator.
Our viewpoint of causality focuses on $P(x_{B}| \sigma(\tilde{x}_A))$ to
capture how the information of $X_A$ affects the potential outcome of $X_B$,
while Pearl's focuses on $P(x_{B}| do(\tilde{x}_A))$ to
capture how the value of $X_A$ affects $X_B$.
In other words, we say that $X_A$ causes $X_B$ iff
``\emph{changing information on $X_A$ leads to a potential change in $X_B$, while keeping everything else constant}''.
Based on this ground, we actually view the causality as the \emph{information transfer} in our \emph{info} intervention framework.
Our viewpoint is different from that in Pearl's \emph{do} intervention (or the general interventionist causality) framework, under which $X_A$ causes $X_B$ iff
``changing $X_A$ leads to a change in $X_B$, while keeping everything else constant''.

Besides the close relationship to the interventionist causality, our \emph{info} intervention framework also builds linkages to
``Law-like'' causality in  physics and experimental causality in statistics, economics and social sciences.
First, the \emph{info} intervention keeps the causal mechanisms as in ``Law-like'' causality. Second, the \emph{info} intervention
creates the counterfactual variables as in experimental causality. These two features
are not owned by Pearl's \emph{do} intervention, and they allow us to
use the counterfactual variables to carry the information on $X_A$ and transfer this information by the unchanged structural equations.

\section{Causal calculus for info intervention}

The \emph{do} intervention is a standard for studying the causality, since it serves (at least) two purposes: communication and theoretical focus (Pearl, 2009).
In the previous sections, our \emph{info} intervention has shown its ability as a standard for communicating about causal questions.
In this section, we will show our \emph{info} intervention can also
be a standard for theoretically focusing on the causal inference. That is,
the theoretical results established for \emph{info} intervention are applicable to calculate
interventional distributions by using observational distributions,
whenever certain conditions hold in the \emph{info}-causal DAG.
For ease of presentation,
the following abbreviations are used below:
\begin{align*}
P(x_{B}|\sigma(\tilde{x}_{A}))&:=P(X_{B}^{\sigma(\tilde{x}_{A})}=x_{B}) \mbox{ in } G^{\sigma(\tilde{x}_{A})},\\
P(x_{B}|\sigma(\tilde{x}_{A}),x_{C})&:=\frac{P(x_{B}, x_{C}|\sigma(\tilde{x}_{A}))}{P(x_{C}|\sigma(\tilde{x}_{A}))}.
\end{align*}

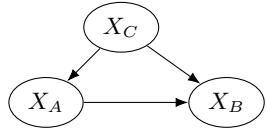
\begin{figure}[htp]
			\begin{tikzpicture}
			\node[state] (x) at (0, 1) {$X_A$};
			\node[state](z) at (1, 2) 	{$X_C$};
			\node[state] (y) at (2.4, 1) {$X_B$};
			\path (z) edge (y);
			\path (z) edge (x);
			\path (x) edge (y);
			\end{tikzpicture}
			\caption{An \emph{info}-causal DAG satisfying the ``back-door'' criterion.}
			\label{fig:back}
	\end{figure}

First, we consider an \emph{info}-causal DAG, in which a set of adjustment variables $X_{C}$ satisfy
the ``back-door'' criterion relative to the ordered pair of variables $(X_A, X_B)$ (see Fig\,\ref{fig:back}).
The ``back-door'' criterion was first given by Pearl (1993), and it is
equivalent to the ignorability condition of Rosenbaum and Rubin (1983).
A DAG satisfying ``back-door'' criterion has been commonly used in treatment-outcome applications, and it enables us
to identify the causal effect of $X_A$ on $X_B$ by adjusting $X_C$.


	\begin{figure}[htp]
			\begin{tikzpicture}
			\node[state] (x) at (0, 1) {$X_A$};
			\node[state] (z) at (1.5, 1)  {$X_{C}$};
			\node[state, dashed](u) at (1.5, 2) 	{$X_D$};
			\node[state] (y) at (3, 1) {$X_B$};
			\path[dashed] (u) edge (y);
			\path[dashed] (u) edge (x);
			\path (x) edge (z);
			\path (z) edge (y);
			\end{tikzpicture}
			\caption{An \emph{info}-causal DAG satisfying the ``front-door'' criterion.}
			\label{fig:front}
	\end{figure}
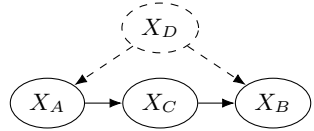

 \begin{thm}[``Back-door'' criterion]
 	For an info-causal DAG $G$ in Fig\,\ref{fig:back},
 	$$
 	P(x_B|\info(\tilde{x}_{A})) = \sum_{x_{C}} P(x_{B}|\tilde{x}_{A}, x_{C}) P(x_{C}).
	$$\label{back_doorthm}
 \end{thm}

When the ``back-door'' criterion holds, Theorem \ref{back_doorthm} shows that
the causal effect of $X_A$ on $X_B$ is identifiable, since $P(x_B|\info(\tilde{x}_{A}))$ (i.e., the
probability of counterfactual variables $X_{B}^{\sigma(\tilde{x}_A)}$) in $G^{\info(\tilde{x}_{A})}$ can be calculated
by using the observational probabilities $P(x_{B}|\tilde{x}_{A}, x_{C})$ and $P(x_{C})$ in $G$.

Second, if no observed variables $X_C$ satisfying ``back-door'' criterion are found in $G$,
we may use an alternative criterion called ``front-door'' criterion. Fig\,\ref{fig:front} plots
an \emph{info}-causal DAG, in which a set of adjustment variables $X_{C}$ satisfy
the ``front-door'' criterion relative to the ordered pair of variables $(X_A, X_B)$ and some hidden variables $X_{D}$.
The ``front-door'' criterion for \emph{do} intervention was studied by Pearl (1995), and below we show that
it can also be used to identify the causal effect for \emph{info}-causal DAG.


\begin{thm}[``Front-door'' criterion]
For an info-causal DAG $G$ in Fig\,\ref{fig:front},
	$$
	P(x_{B}|\info(\tilde{x}_A)) = \sum_{x_{C}}  P(x_{C}|\tilde{x}_{A}) \sum_{x_A} P(x_{B}|x_{C}, x_{A}) P(x_{A}).
	$$\label{front_doorthm}
\end{thm}

When  the ``front-door'' criterion holds, Theorem \ref{front_doorthm} shows that
the causal effect of $X_A$ on $X_B$ is identifiable, since $P(x_B|\info(\tilde{x}_{A}))$  in $G^{\info(\tilde{x}_{A})}$  can be calculated
by using the observational probabilities $P(x_{C}|\tilde{x}_{A})$, $P(x_{B}|x_{C}, x_{A})$ and $P(x_{A})$ in $G$.

Third, to deal with more complicated DAGs beyond ``back-door'' and ``front-door'' criteria,
Pearl (1995) provided 3 rules for \emph{do} intervention,
which enable us to identify a causal query and turn a causal question into a statistical estimation problem.
Specifically, Pearl's 3 rules describe graphical criteria for
\begin{enumerate} 
	\item insertion/deletion of observations,
	\item action/observation exchange,
	\item insertion/deletion of actions.
\end{enumerate}
By using these three rules, the expression of \emph{do} intervention probability may be reduced step-wisely to
an equivalent expression involving only observational probabilities.

Denote by $\indep_d$ the $d$-separation, $anc(I)$ the ancestor nodes of every node in $I$, and $G_{\bar{I}}$ the graph obtained by deleting from $G$ all arrows pointing to nodes in $I$.
Similar to Pearl's 3 rules for $do$ intervention, we can present our three rules for \emph{info} intervention.

\begin{thm}[Three rules for \emph{info} intervention]
	\label{thm:rules}
	For an info-causal DAG $G$, $A, B, C$ and $D$ are its arbitrary disjoint node sets. Then,
	
	\textbf{Rule 1} (Insertion/deletion of observations) \\
	$P(x_{B}|\info(\tilde{x}_{A}), x_{C}, x_{D}) = P(x_{B}|\info(\tilde{x}_A), x_{D})$ if $B \indep_d C|D$ in $G^{\info(\tilde{x}_{A})}$;
	
	\textbf{Rule 2} (Action/observation exchange) \\
	$P(x_{B}|\info(\tilde{x}_{A}), \info(\tilde{x}_{C}), x_{D}) = P(x_{B}|\info(\tilde{x}_A), \tilde{x}_{C}, x_{D})$ if $B \indep_d C|D$ in $G^{\info(\tilde{x}_A, \tilde{x}_C)}$;
	
	\textbf{Rule 3} (Insertion/deletion of actions) \\
	$P(x_{B}|\info(\tilde{x}_A), \info(\tilde{x}_{C}), x_{D}) = P(x_{B}|\info(\tilde{x}_{A}), x_{D})$ if  $B \indep_d C|D$
in $G^{\info(\tilde{x}_A)}_{\overline{C/anc(D)}}$,

\noindent where $C/anc(D)$ is the set of $C$-nodes that are not ancestors of any $D$-node.
\end{thm}



In some applications, we may use the following simpler version of three rules in Theorem \ref{thm:rules}.

\begin{thm}
	\label{thm:rules:simple}
	For an info-causal DAG $G$, $A, B, C$ and $D$ are its arbitrary disjoint node sets. Then,
	
	\textbf{Rule 1} (Insertion/deletion of observations) \\
	$P(x_{B}|\info(\tilde{x}_{A}), x_{C}, x_{D}) = P(x_{B}|\info(\tilde{x}_A), x_{D})$ if $B \indep_d C|D$ in $G^{\info(\tilde{x}_{A})}$;
	
	\textbf{Rule 2} (Action/observation exchange) \\
	$P(x_B|\info(\tilde{x}_A), x_C) = P(x_B|\tilde{x}_A, x_C)$  if $B\indep_d A|C$ in $G^{\info(\tilde{x}_A)}$;
	
	\textbf{Rule 3} (Insertion/deletion of actions) \\
	$P(x_B|\info(\tilde{x}_A)) = P(x_B)$ if there are no causal paths from $A$ to $B$ in $G$.
\end{thm}

In view of Theorems \ref{back_doorthm}--\ref{front_doorthm}, we can see that our formulas
on ``back-door''/``front-door'' criteria are the same as Pearl's formulas in Pearl (1995). This 
is because the distributions of all non-intervention variables are the same
in both \emph{info}- and \emph{do}-intervention DAGs, and the random intervention variables in \emph{info}-intervention DAG
behave similarly as the deterministic intervention variables in \emph{do}-intervention DAG, due to the fact that
the intervention variables in \emph{info}-intervention DAG with only possible converging arrows do not cause any other variables.
Indeed, by (\ref{eq:sigma:do}) and (\ref{eq:sigma}), it is straightforward to see that for arbitrary disjoint node sets $A, B$ and $C$ in $V$,
\begin{align}\label{complete_rule}
P(x_B|do(\tilde{x}_A), x_C) = P(x_B|\info(\tilde{x}_A), x_C).
\end{align}
The result (\ref{complete_rule}) implies that our formulas on three Rules in Theorem \ref{thm:rules}
are also the same as Pearl's formulas on three Rules in Pearl (1995).
Therefore, since Pearl's causal calculus based on \emph{do} intervention is complete (Huang and Valtorta, 2012), our causal calculus based on \emph{info} intervention is also complete, meaning that if a causal effect $P(x_B|\sigma(\tilde{x}_A), x_C)$ is identifiable, it can be calculated
by using a sequence of Rules 1--3 in Theorem \ref{thm:rules}.

The result (\ref{complete_rule}) also indicates that Pearl's causal calculus and our causal calculus are exchangeable, but
this does not mean the same manipulating convenience in both frameworks. Theorem \ref{Equivalence} below shows that
our conditions for checking Rules 1--3 in Theorem \ref{thm:rules} are equivalent to those
for checking Rules 1--3 in Pearl (1995), and they tend to be more convenient for use since
the intervention nodes $A$ are not involved as part of conditioning set in \emph{info}-intervention DAG.

\begin{thm}[Equivalence of checking conditions]
	\label{Equivalence}
	For an info-causal DAG $G$, $A, B, C$ and $D$ are its arbitrary disjoint node sets. Then,
	
	(i) $B \indep_d C|D$ in $G^{\info(\tilde{x}_{A})}$ $\Longleftrightarrow$ $B \indep_d C|A, D$ in $G_{\overline{A}}$;
	
	(ii) $B \indep_d C|D$ in $G^{\info(\tilde{x}_A, \tilde{x}_C)}$ $\Longleftrightarrow$ $B \indep_d C|A, D$ in $G_{\overline{A}}^{\sigma(\tilde{x}_C)}$;
	
	(iii) $B \indep_d C|D$ in $G^{\info(\tilde{x}_A)}_{\overline{C/anc(D)}}$ $\Longleftrightarrow$  $B \indep_d C|A, D$ in $G_{\overline{A}, \overline{C/anc(D)}}$.
\end{thm}


Denote $X_A\indep X_B|X_C$ by the conditional independence of
$X_A$ and $X_B$, given $X_C$. To end this section, we re-visit an example in Richardson and Robins (2013).



 \begin{figure}[htp]
 \begin{subfigure}[tb]{0.35\textwidth}
 	\centering
 	\begin{tikzpicture}
 	\node[state] (a) at (-0.5, 1) {$X_{A_1}$};
 	\node[state] (z) at (1, 1)  {$X_C$};
 	\node[state](b) at (1, 2) 	{$X_{A_2}$};
 	\node[state, dashed] (h) at (-0.5, 2) {$X_D$};
 	\node[state] (y) at (2.2, 1.5) {$X_B$};
 	\path (h) edge (z);
 	\path (h) edge (b);
 	\path (b) edge (y);
 	\path (a) edge (z);
 	\path (z) edge (b);
 	\path (z) edge (y);
 	\end{tikzpicture}
 	\caption{A DAG $G$.}
    \label{fig:indep1}
 \end{subfigure}
 \hfill
 \begin{subfigure}[tb]{0.55\textwidth}
 	\centering
 	\begin{tikzpicture}
 	\node[state] (a) at (-0.5, 1) {$X_{A_1}$};
 	\node[state] (z) at (2, 1)  {$X_{C}^{\info(\tilde{x}_{A_1})}$};
 	\node[state](b) at (2, 3) 	{$X_{A_2}^{\info(\tilde{x}_{A_1})}$};
 	\node[state, dashed] (h) at (-0.5, 3) {$X_{D}$};
 	\node[state] (y) at (4, 2) {$X_{B}^{\info(\tilde{x}_{A_1}, \tilde{x}_{A_2})}$};
 	\path (h) edge (z);
 	\path (h) edge (b);
 	\path (z) edge (b);
 	\path (z) edge (y);
 	\end{tikzpicture}
 	\caption{$G^{\info(\tilde{x}_{A_1}, \tilde{x}_{A_2})}$.}
 	\label{fig:indep3}
 \end{subfigure}
 \caption{A DAG $G$ and its \emph{info}-intervention DAG.}
 \label{fig:indep}
\end{figure}
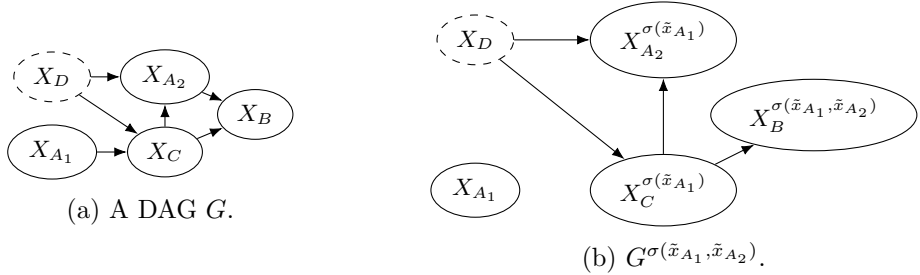
 \begin{example}
 	\label{eg:info_causal_DAG}
 	Consider a DAG $G$ in Fig\,\ref{fig:indep}(a), where $i\not\in desc(A_1)$ for any $i\in A_1$, and
 $i\not\in desc(A_2)$ for any $i\in A_2$. Fig\,\ref{fig:indep}(b) plots the info-intervention DAG $G^{\info(\tilde{x}_{A_1}, \tilde{x}_{A_2})}$, where we have used the fact
 \begin{align*}
X_{A_1}^{\info(\tilde{x}_{A_1}, \tilde{x}_{A_2})}&=\big(X_{A_1}^{\info(\tilde{x}_{A_1})}\big)^{\info(\tilde{x}_{A_2})}=
\big(X_{A_1}\big)^{\info(\tilde{x}_{A_2})}=X_{A_1},\\
X_{A_2}^{\info(\tilde{x}_{A_1}, \tilde{x}_{A_2})}&=\big(X_{A_2}^{\info(\tilde{x}_{A_2})}\big)^{\info(\tilde{x}_{A_1})}
=X_{A_2}^{\info(\tilde{x}_{A_1})},\\
X_{C}^{\info(\tilde{x}_{A_1}, \tilde{x}_{A_2})}&=\big(X_{C}^{\info(\tilde{x}_{A_1})}\big)^{\info(\tilde{x}_{A_2})}=X_{C}^{\info(\tilde{x}_{A_1})},\,\,
X_{D}^{\info(\tilde{x}_{A_1}, \tilde{x}_{A_2})}=X_D.
\end{align*}
Then,
 $X_{B}^{\info(\tilde{x}_{A_1}, \tilde{x}_{A_2})} \indep X_{A_2}^{\info(\tilde{x}_{A_1})} | X_{A_1}, X_{C}^{\info(\tilde{x}_{A_1})}$, since  $B \indep_d A_2 | A_1, C$ in Fig\,\ref{fig:indep}(b). Note that this conclusion was also proved in Richardson and Robins (2013) by  constructing a single-world intervention graph (SWIG).

Besides the checking of independence between counterfactual variables, we can also calculate $P(x_B|\info(\tilde{x}_{A_1}, \tilde{x}_{A_2}), x_C)$
(i.e.,
the conditional
probability of counterfactual variables $X_{B}^{\info(\tilde{x}_{A_1}, \tilde{x}_{A_2})}$ given $X_{C}^{\info(\tilde{x}_{A_1}, \tilde{x}_{A_2})}$) by
\begin{align*}
   &P(x_B|\info(\tilde{x}_{A_1}, \tilde{x}_{A_2}), x_C) \\
   &= P(x_B|\info(\tilde{x}_{A_1}), \tilde{x}_{A_2}, x_C) \,\,\, \mbox{(by Rule 2 in Theorem }\ref{thm:rules}) \\
    &= P(x_B|\tilde{x}_{A_2}, x_C) \,\,\, \mbox{(by Rule 3 in Theorem }\ref{thm:rules}).
\end{align*}
Note that Richardson and Robins (2013) assumed the modularity condition
$$P(x_B|\info(\tilde{x}_{A_1}, \tilde{x}_{A_2}), x_C)=P(x_B|\tilde{x}_{A_1}, \tilde{x}_{A_2}, x_C)$$
to facilitate the causal calculus in SWIG. This modularity condition, which links the counterfactual probability
 to the observational probability, is not needed in our info intervention framework.
 \end{example}

\section{Extension to generalized info intervention} In our \emph{info} intervention framework,
the intervention variables with no output edges in the \emph{info}-intervention graph, have no further causal effects
on other variables in the graph. In some applications, this phenomenon may not be desirable in terms of communication and
theoretical focus.
To tackle this problem, we extend our \emph{info} intervention to the generalized \emph{info} intervention in this section, and
this extension seems challenging for the \emph{do} intervention.

\subsection{Generalized info intervention}
Consider an SCM $M=(G^+, \mathcal{X}, P, f)$ for $X_V$.
For $j, k\in V$, define  an information function
$\sigma_{jk}: \mathcal{X}_{j}\to \mathcal{X}_{j}$ to capture the transferred information on $X_j$
from node $j$ to node $k$. For example, we have
$\sigma_{jk}(X_j)=\tilde{x}_{j}$ for $k\in ch(j)$ in the \emph{info} intervention framework, meaning that
all the child nodes of $j$ have received the information  $X_j=\tilde{x}_j$.
Therefore, by choosing different formulas of $\sigma_{jk}$, we can send out different information on
$X_j$ from node $j$ to node $k$, leading to different types of information intervention.

As the information intervention is uniquely determined by the information function
$\sigma_{jk}$, we can collect several information functions to form an information set
\begin{align}\label{sigma_F}
\sigma(F):=\big\{\sigma_{jk}: (j, k)\in F\subseteq V^2\big\},
\end{align}
where $F$ is the information index set.
We call $\sigma(F)$ the generalized \emph{info} intervention operator, since it
includes different information interventions in the graph. Below, we
show how $\sigma(F)$ manipulates in the SCM.

\begin{definition}[Generalized \emph{info} intervention]
	Given an SCM $M=(G^+, \mathcal{X}, P, f)$ for $X_V$, the generalized info intervention
$\sigma(F)$ in (\ref{sigma_F}) maps $M$ to the generalized info-intervention model $M^{\info(F)}=(G^+,\mathcal{X}, P, f)$ for $X^{\sigma(F)}_V$, where
    $$
    X^{\sigma(F)}_v= {f}_v(\widetilde{X}_{V \cap pa(v)}, X_{U \cap pa(v)})
    $$
    with $\widetilde{X}_{V \cap pa(v)}:=\{\widetilde{X}_{jv}\}_{j\in pa(v)}$ such that
    $\widetilde{X}_{jv} = \sigma_{jv}(X_j)$ if $(j, v) \in F$ else $X^{\sigma(F)}_j$.\label{gii_def}
\end{definition}

Based on Definition \ref{gii_def},
we can first show that for two disjoint information index sets $F_1$ and $F_2$,
$X^{\info(F_1, F_2)}_v:=\big(X^{\info(F_1)}_v\big)^{\info(F_2)}$ has the
commutative property, that is, $X^{\info(F_1, F_2)}_v=X^{\info(F_2, F_1)}_v$ for all $v\in V$.

Second, we know that $M^{\info(F)}$ does not delete any structural equations $f_V$ from the model, but just sends out the information
$X_{j}$ is replaced by $\widetilde{X}_{jk}$ to the node $k$ if $(j,k)\in F$, and the variable at node $k$ becomes
a counterfactual variable with a hypothetical input $X_{j}=\widetilde{X}_{jk}$ in its structural equation.
When $\sigma_{jk}(\cdot)$ is a constant function
(as in the \emph{info} intervention), the value of $X_j$ is completely known to
the node $k$ after intervention, and hence the edge from $j$ to $k$ (denoted by $j\to k$) is removed in $M^{\info(F)}$.
 When $\sigma_{jk}(\cdot)$ is not a constant function, the value of $X_j$ is still unknown to
 the node $k$ after intervention, and the edge $j\to k$ is replaced by the information edge (denoted by $j\overset{\sigma_{jk}}{\to} k$)
 to transfer the information $X_{j}$ is replaced by $\widetilde{X}_{jk}$ in $M^{\info(F)}$.
In both cases, the information is further transferred to $desc(k)$, making the variables at $desc(k)$
become the counterfactual variables.

Third, it is interesting to see that the generalized \emph{info} intervention nests several information interventions as its special cases:
\begin{enumerate} 
	\setlength{\itemsep}{-1pt}
	\item \emph{Info} intervention
	\begin{itemize}
		\setlength{\itemsep}{0pt}		
		\item Information function: $\sigma_{jk}(X_j)=\tilde{x}_{j}$ for $j\in J$ and $k\in ch(J)$.
		\item Transferred information: All the child nodes of $J$ receive the information $X_j$ is replaced by $\tilde{x}_j$.
		\item Intervention graph: All output edges of nodes $J$ are deleted.
	\end{itemize}
	\item Precision \emph{Info} intervention
	\begin{itemize}
		\setlength{\itemsep}{0pt}		
		\item Information function: $\sigma_{jk}(X_j)=\tilde{x}_{j}$ for $j\in J$ and $k\in K\subseteq ch(J)$.
		\item Transferred information: Only the nodes $K$ receive the information $X_j$ is replaced by $\tilde{x}_j$.
		\item Intervention graph: All output edges from nodes $J$ to nodes $K$ are deleted.
	\end{itemize}
	\item Shift \emph{Info} intervention
	\begin{itemize}
		\setlength{\itemsep}{0pt}		
		\item Information function: $\sigma_{jk}(X_j)=X_j+\tilde{s}_{j}$ for $j\in J$ and $k\in K\subseteq ch(J)$.
		\item Transferred information: Only the nodes $K$ receive the information $X_j$ is replaced by $X_j+\tilde{s}_{j}$.
		\item Intervention graph: Each edge $j\to k$ is replaced by $j\overset{\sigma_{jk}}{\to} k$.
    \end{itemize}
    \item Transform \emph{Info} intervention
	\begin{itemize}
		\setlength{\itemsep}{0pt}		
		\item Information function: $\sigma_{jk}(X_j)=g_{jk}(X_j)$ for $j\in J$, $k\in K\subseteq ch(J)$, and a given function $g_{jk}(\cdot)$.
		\item Transferred information: Each node $k$ receives the information $X_{j}$ is replaced by $g_{jk}(X_j)$.
		\item Intervention graph:  Each edge $j\to k$ is replaced by $j\overset{\sigma_{jk}}{\to} k$ if $g_{jk}(\cdot)$ is not a constant function, else it is deleted.
	\end{itemize}
\end{enumerate}

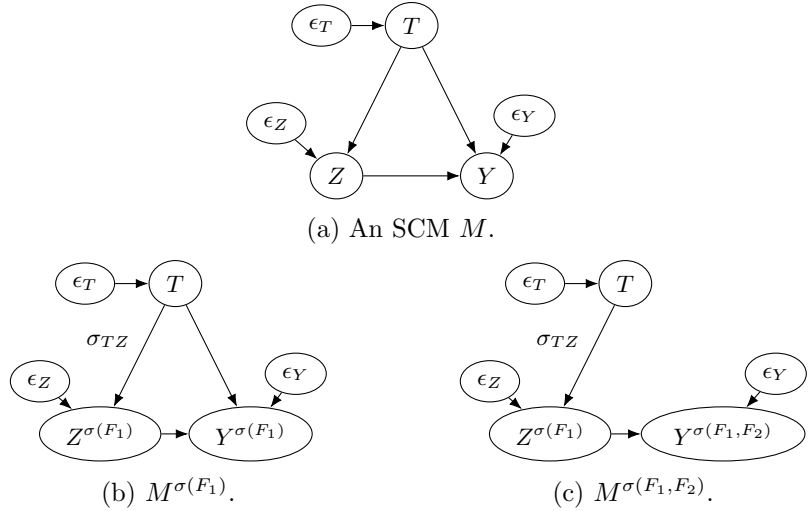
\begin{figure}[htp]
	\begin{subfigure}[tb]{1.0\textwidth}
    \centering
    \begin{tikzpicture}
    \node[state] (t) at (1, 2) {$T$};
    \node[state] (z) at (0, 0) {$Z$};
    \node[state] (y) at (2, 0) {$Y$};
    \node[state] (te) at (-0.2, 2) {$\epsilon_T$};
    \node[state] (ze) at (-0.8, 0.7) {$\epsilon_Z$};
    \node[state] (ye) at (2.5, 0.8) {$\epsilon_Y$};
    \path (t) edge (z);
    \path (t) edge (y);
    \path (z) edge (y);
    \path (te) edge (t);
    \path (ze) edge (z);
    \path (ye) edge (y);
    \end{tikzpicture}
    \caption{An SCM $M$.}
	\end{subfigure}
	\vfill
	\begin{subfigure}[tb]{0.48\textwidth}
    \centering
    \begin{tikzpicture}
    \node[state] (t) at (1, 2) {$T$};
    \node[state] (z) at (0, 0) {$Z^{\sigma(F_1)}$};
    \node[state] (y) at (2, 0) {$Y^{\sigma(F_1)}$};
    \node[state] (te) at (-0.2, 2) {$\epsilon_T$};
    \node[state] (ze) at (-0.8, 0.7) {$\epsilon_Z$};
    \node[state] (ye) at (2.6, 0.8) {$\epsilon_Y$};
    \path (t) edge[above left=5]   node {$\sigma_{TZ}$} (z);
    \path (t) edge (y);
    \path (z) edge (y);
    \path (te) edge (t);
    \path (ze) edge (z);
    \path (ye) edge (y);
    \end{tikzpicture}
    \caption{$M^{\info(F_1)}$.}
	\end{subfigure}
	\begin{subfigure}[tb]{0.48\textwidth}
    \centering
    \begin{tikzpicture}
    \node[state] (t) at (1, 2) {$T$};
    \node[state] (z) at (0, 0) {$Z^{\sigma(F_1)}$};
    \node[state] (y) at (2.3, 0) {$Y^{\sigma(F_1, F_2)}$};
    \node[state] (te) at (-0.2, 2) {$\epsilon_T$};
    \node[state] (ze) at (-0.8, 0.7) {$\epsilon_Z$};
    \node[state] (ye) at (3, 0.8) {$\epsilon_Y$};
    \path (t) edge[above left=5]   node {$\sigma_{TZ}$} (z);
    \path (z) edge (y);
    \path (te) edge (t);
    \path (ze) edge (z);
    \path (ye) edge (y);
    \end{tikzpicture}
    \caption{$M^{\info(F_1, F_2)}$.}
	\end{subfigure}
	\caption{An SCM and its two generalized \emph{info}-intervention SCMs.}
	\label{fig:generalized_example}
\end{figure}

Clearly, the transform \emph{info} intervention in Case 4 nests other interventions in Cases 1-3, and all interventions above could be used together. To further illustrate how to manipulate $\sigma(F)$, we consider the following example:

\begin{example}
	\label{eg4}
	Consider an SCM $M$ with three univariate endogenous variables $T$, $Z$ and $Y$,
	and three latent variables $\epsilon_T, \epsilon_Z$ and  $\epsilon_Y$ in Fig\,\ref{fig:generalized_example}(a). The  structural equations
    for $T$, $Z$ and $Y$ are given by
	$$
	\begin{aligned}
	\begin{cases}
	T = f_T(\epsilon_T),  \\   
	Z = f_Z(T, \epsilon_Z), \\
	Y = f_Y(T, Z, \epsilon_Y).
	\end{cases}
	\end{aligned}
	$$
   For this SCM, we consider two generalized info interventions $\sigma(F_1)$ and $\sigma(F_1, F_2)$, where
   $F_1=\{\sigma_{TZ}\}$ and $F_2=\{\sigma_{TY}\}$ with $\sigma_{TZ}(T)=T+\tilde{s}$ and $\sigma_{TY}(T)=\tilde{t}$.
Based on Definition \ref{gii_def}, the
	generalized info-intervention SCM $M^{\sigma(F_1)}$ given in Fig\,\ref{fig:generalized_example}(b) has the following structural equations:
	$$
	\begin{aligned}
	\begin{cases}
	T = f_T(\epsilon_T),  \\
	Z^{\sigma(F_1)} = f_Z(T+\tilde{s}, \epsilon_Z), \\
	Y^{\sigma(F_1)} = f_Y(T, Z^{\sigma(F_1)}, \epsilon_Y),
	\end{cases}
	\end{aligned}
	$$    	
where we have used the fact that $T^{\sigma(F_1)}=T$.
Next, the generalized info-intervention SCM $M^{\sigma(F_1, F_2)}$ given in Fig\,\ref{fig:generalized_example}(c) has the following structural equations:
	$$
	\begin{aligned}
	\begin{cases}
	T = f_T(\epsilon_T),  \\
	Z^{\sigma(F_1)} = f_Z(T+\tilde{s}, \epsilon_Z), \\
	Y^{\sigma(F_1, F_2)} = f_Y(\tilde{t}, Z^{\sigma(F_1)}, \epsilon_Y),
	\end{cases}
	\end{aligned}
	$$
where we have used the fact that $T^{\sigma(F_1, F_2)}=T$ and $Z^{\sigma(F_1, F_2)}=Z^{\sigma(F_1)}$.
Note that the causal mechanisms (i.e., $f_{Z}$, $f_{T}$ and $f_{Y}$)
are unchanged in $M^{\info(F_1)}$ and $M^{\sigma(F_1, F_2)}$, while $Z$ and $Y$ in $M$ become the counterfactual variables $Z^{\sigma(F_1)}$
 and $Y^{\sigma(F_1, F_2)}$ in $M^{\sigma(F_1, F_2)}$, respectively.
Moreover, from Fig\,\ref{fig:generalized_example}(c), the $d$-separation argument implies  that given $Z^{\sigma(F_1)}$,   $T$ and $Y^{\sigma(F_1, F_2)}$ are independent.
\end{example}

In terms of communication, the generalized \emph{info} intervention allows us to answer more complicated
causal questions than the \emph{info} intervention. For example, we could use the precision \emph{info} intervention
to study how the intervention variables affect other variables,
apply the shift \emph{info} intervention to answer the counterfactual question: What if the public had received the information
the price of certain brand increases (or decreases) by 1 dollar? or manipulate the transform \emph{info} intervention to answer the counterfactual question:  What if the public  had received the information
the price of certain brand increases (or decreases) by 10\%?

\subsection{Generalized info-causal DAG} Analogous to the \emph{info}-causal DAG in Definition \ref{def_4_1},
we can define the generalized \emph{info}-causal DAG in the generalized information intervention framework.

\begin{definition}[Generalized \emph{info}-causal DAG]
	Consider a DAG $G = (V, E)$ and a random vector $X$ with distribution $P$. Then, $G$ is called a generalized info-causal DAG for $X$ if $P$ satisfies the following:
	\begin{enumerate}
		\setlength{\itemsep}{0pt}
		\item $P$ factorizes, and thus is Markov, according to $G$,
		\item for any $F \subseteq V^2$ and any $e_{jk}:=\sigma_{jk}(x_j)$ in the domain of $X_j$,
		\begin{equation}\label{eq:sigma_generalized}
			P(x|\info(F)) = \prod_{k \in V} P(x_k|x_{pa(k)}^*),
		\end{equation}		
		where $x_{pa(k)}^*:=\{x_{jk}^{*}\}_{j\in pa(k)}$ such that
$x_{jk}^*=x_j$ if $(j, k) \notin F$ else $e_{jk}$.
	\end{enumerate}
\end{definition}

Comparing (\ref{eq:sigma}) with (\ref{eq:sigma_generalized}), the only difference is
that $P(x_k|x_{pa(k)}^*)$ replaces
$x_{jk}^{*}\in x_{pa(k)}^*$ with $(j, k)\in F$, by $\tilde{x}_j$ in the \emph{info}-causal DAG, while this $x_{jk}^{*}$ is replaced by  $e_{jk}$ in the generalized \emph{info}-causal DAG.

Next, similar to the \emph{info}-intervention DAG in Definition \ref{info_inter_dag}, we can define the generalized \emph{info}-intervention DAG as follows:

\begin{definition}[Generalized \emph{info}-intervention DAG]
	Consider a generalized info-causal DAG $G = (V, E)$ for a random vector $X$, and its generalized info intervention $\sigma(F)$.
Denote a node set $A_{F}=\{i: (i, j)\in F\}$.
The generalized info-intervention DAG, denoted by $G^{\sigma(F)}$, is for $X^{\sigma(F)}$, which has the generalized info-intervention distribution $P(x| \sigma(F))$ in (\ref{eq:sigma_generalized}), where $X^{\sigma(F)}$ is defined in the same way as $X$, except that the variables at descendant nodes of $A_{F}$ (say, $X_{desc(A_F)}$) are replaced
by the counterfactual variables (say, $X_{desc(A_F)}^{\sigma(F)}$).
\end{definition}

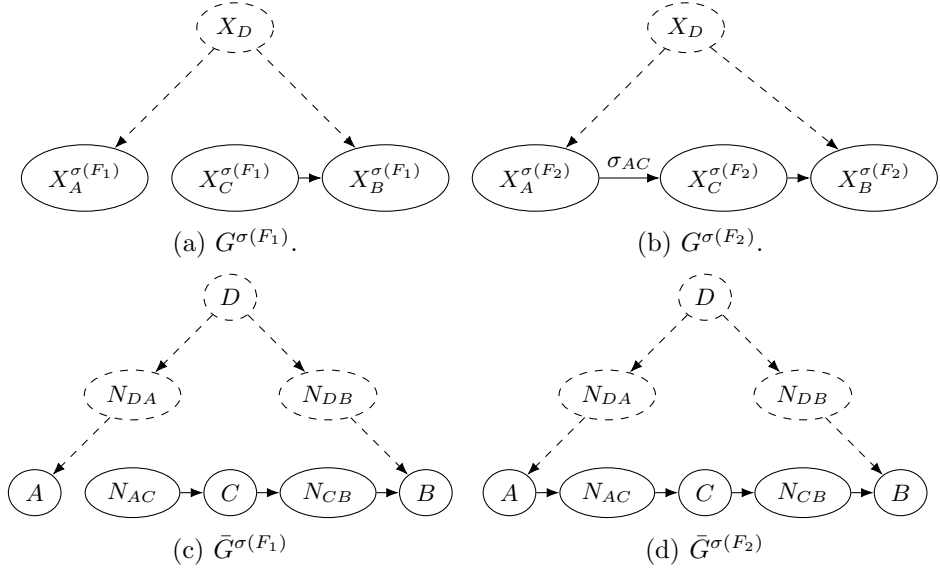
\begin{figure}[htp]
    \vfill
	\begin{subfigure}[tb]{0.48\textwidth}
    \centering
    \begin{tikzpicture}
    \node[state, dashed] (d) at (2, 2) {$X_D$};
    \node[state] (a) at (0, 0) {$X_A^{\sigma(F_1)}$};
    \node[state] (c) at (4, 0) {$X_B^{\sigma(F_1)}$};
    \node[state] (b) at (2, 0) {$X_C^{\sigma(F_1)}$};
    \path[dashed] (d) edge (a);
    \path[dashed] (d) edge (c);
    \path (b) edge (c);
    \end{tikzpicture}
    \caption{$G^{\sigma(F_1)}$.}
	\end{subfigure}
	\begin{subfigure}[tb]{0.48\textwidth}
    \centering
    \begin{tikzpicture}
    \node[state, dashed] (d) at (2, 2) {$X_D$};
    \node[state] (a) at (0, 0) {$X_A^{\sigma(F_2)}$};
    \node[state] (c) at (4.5, 0) {$X_B^{\sigma(F_2)}$};
    \node[state] (b) at (2.5, 0) {$X_C^{\sigma(F_2)}$};
    \path (d)[dashed] edge (a);
    \path (d)[dashed] edge (c);
    \path (a) edge node {$\info_{AC}$} (b);
    \path (b) edge (c);
    \end{tikzpicture}
    \caption{$G^{\sigma(F_2)}$.}
	\end{subfigure}	
	\vfill
	\begin{subfigure}[tb]{0.49\textwidth}
    \centering
    \begin{tikzpicture}
    \node[state, dashed] (d) at (2.6, 2.6) {$D$};
    \node[state] (a) at (0, 0) {$A$};
    \node[state, dashed] (da) at (1.3, 1.3) {$N_{DA}$};
    \node[state] (b) at (5.2, 0) {$B$};
    \node[state] (c) at (2.6, 0) {$C$};
    \node[state] (ac) at (1.3, 0) {$N_{AC}$};
    \node[state] (cb) at (3.9, 0) {$N_{CB}$};
    \node[state, dashed] (db) at (3.9, 1.3) {$N_{DB}$};
    \path (d)[dashed] edge (da);
    \path (da)[dashed] edge (a);
    \path (d)[dashed] edge (db);
    \path (db)[dashed] edge (b);
    \path (ac) edge (c);
    \path (c) edge (cb);
    \path (cb) edge (b);
    \end{tikzpicture}
    \caption{$\bar{G}^{\info(F_1)}$}
	\end{subfigure}
	\begin{subfigure}[tb]{0.49\textwidth}
    \centering
    \begin{tikzpicture}
    \node[state, dashed] (d) at (2.6, 2.6) {$D$};
    \node[state] (a) at (0, 0) {$A$};
    \node[state, dashed] (da) at (1.3, 1.3) {$N_{DA}$};
    \node[state] (b) at (5.2, 0) {$B$};
    \node[state] (c) at (2.6, 0) {$C$};
    \node[state] (ac) at (1.3, 0) {$N_{AC}$};
    \node[state] (cb) at (3.9, 0) {$N_{CB}$};
    \node[state, dashed] (db) at (3.9, 1.3) {$N_{DB}$};
    \path (d)[dashed] edge (da);
    \path (da)[dashed] edge (a);
    \path (d)[dashed] edge (db);
    \path (db)[dashed] edge (b);
    \path (ac) edge (c);
    \path (c) edge (cb);
    \path (cb) edge (b);
    \path (a) edge (ac);
    \end{tikzpicture}
    \caption{$\bar{G}^{\info(F_2)}$}
	\end{subfigure}
	\caption{Two generalized \emph{info}-intervention DAGs for the DAG in Fig\,\ref{fig:front}, and their related augmented DAGs.
For $G^{\sigma(F_1)}$, $F_1=\{\sigma_{AC}\}$ and $\sigma_{AC}(X_A)=\tilde{x}_A$ (i.e., $\sigma_{AC}(\cdot)$ is a constant function).
For $G^{\sigma(F_2)}$, $F_2=\{\sigma_{AC}\}$ and $\sigma_{AC}(\cdot)$ is not a constant function.}
	\label{fig:last_pic}
\end{figure}

For $(j, k) \in F$, we know that $e_{jk}$ in $G^{\sigma(F)}$ still depends on $x_j$ unless $\sigma_{jk}(\cdot)$ is a constant function. Therefore, when $\sigma_{jk}(\cdot)$ is not a constant function, the edge $j\to k$ can not be deleted as in the \emph{info}-intervention DAG, but need be replaced by the information edge
$j\overset{\sigma_{jk}}{\to} k$ to process the information transfer in $G^{\sigma(F)}$. To see this difference clearly, Fig\,\ref{fig:last_pic}(a)-(b)
plot two generalized \emph{info}-intervention DAGs $G^{\sigma(F_1)}$ and $G^{\sigma(F_2)}$ for the DAG $G$ satisfying ``front-door'' criterion in Fig\,\ref{fig:front}.
In $G^{\sigma(F_1)}$, since the information function $\sigma_{AC}(\cdot)$ is a constant function,
the edge $A\to C$ is deleted. On the contrary, the edge $A\to C$ is replaced by the information edge $A\overset{\sigma_{AC}}{\to} C$ in $G^{\sigma(F_2)}$,
since the related information function $\sigma_{AC}(\cdot)$ is not a constant function.

As before, we can similarly define the causal effect of $X_A$ on $X_B$ and the causal identifiability in the generalized \emph{info}-causal DAG, and the details are omitted for saving space.  For the theoretical focus, the generalized \emph{info}-causal DAG can still have Rules 1 and 3 for causal calculus.
To justify the validity of Rules 1 and 3, we need check the $d$-separation conditions in the augmented DAG and its generalized intervention DAG.

\begin{definition}[Augmented DAG \& Intervention augmented DAG]
Consider a DAG $G = (V, E)$ and a generalized info intervention $\sigma(F)$. Then, its augmented DAG
is $\bar{G}=(\bar{V}, \bar{E})$, where $\bar{G}$ is constructed by inserting a new node $N_{ij}$ in
every edge $i\to j$ in $G$. Moreover, the generalized intervention augmented DAG
for $\sigma(F)$ is
$\bar{G}^{\sigma(F)}=(\bar{V}^{\sigma(F)}, \bar{E}^{\sigma(F)})$, where
$\bar{V}^{\sigma(F)}=\bar{V}$ and $\bar{E}^{\sigma(F)}=\bar{E}/E_F$ with
\begin{align*}
E_F&=\{\mbox{the edge } i\to N_{ij} : (i, j)\in F \mbox{ and } \sigma_{ij}(\cdot)= \mbox{a constant}\}.
\end{align*}
Here, $N_{ij}\in \bar{V}$ is called the information node. \label{aug_dag}
\end{definition}

By Definition \ref{aug_dag}, the augmented DAG $\bar{G}^{\sigma(F)}$ essentially plugs in
a new information node $N_{ij}$ between node $i$ and node $j$ in $G$, and the edge $i\to N_{ij}$ exists only when $\sigma_{ij}(\cdot)$ is not a constant
function. To see this point clearly, we plot two augmented DAGs $\bar{G}^{\sigma(F_1)}$ and $\bar{G}^{\sigma(F_2)}$ in Fig\,\ref{fig:last_pic}(c)-(d), from which we can see that four new nodes $N_{AC}$, $N_{CB}$, $N_{DA}$ and $N_{DB}$ are added into the graph, and the edge $A\to N_{AC}$ is deleted in $\bar{G}^{\sigma(F_1)}$.

We are now ready to give Rules 1 and 3 in the generalized \emph{info} intervention framework.

\begin{thm}
	\label{thm:gg_3_rules}
	For a generalized info-causal DAG $G=(V, E)$, $B$, $C$ and $D$ are its arbitrary disjoint node sets.
Consider two generalized info interventions $\sigma(F_1)$ and $\sigma(F_2)$ such that
$N_{F_1}\cap N_{F_2}=\emptyset$. Then,
	
	\textbf{Rule 1} (Insertion/deletion of observations) \\
	$P(x_{B}|\info(F_1), x_{C}, x_{D}) = P(x_B|\info(F_1), x_D)$ if $B \indep_d C|D$ in $\bar{G}^{\info(F_1)}$;
	
	\textbf{Rule 3} (Insertion/deletion of actions) \\
	$P(x_B|\info(F_1), \info(F_2), x_D) = P(x_B|\info(F_1), x_D)$ if  $B\indep_d N_{F_2}|D$ in $\bar{G}^{\info(F_1)}_{\overline{N_{F_2}/anc(D)}}$.
\end{thm}

%

The proofs of Theorem \ref{thm:gg_3_rules} are omitted, since they are similar to those of
Theorem \ref{thm:rules}.
Theorem \ref{thm:gg_3_rules} shows that we only need check the $d$-separation condition in the augmented DAG to
use Rules 1 and 3.  As an application, we use Rule 3 in this theorem to
identify the causal effect under  ``front-door'' criterion.

\begin{example}
Consider a generalized info-causal DAG $G$ in Fig\,\ref{fig:front}, and
let $\sigma(F)$ be its generalized info intervention, where
$F=\{\sigma_{AC}\}$ and $\sigma_{AC}(X_A)$ $=g_{AC}(X_A)$ for a given function $g_{AC}(\cdot)$.
In this case, we have
\begin{align*}
P(x_B|\info(F)) &= \sum_{x_{C}} \sum_{x_{A}} P(x_B|x_C, x_A, \info(F) ) P(x_C, x_A|\info(F)) \\
&= \sum_{x_C}\sum_{x_A} P(x_B|x_C, x_A) P(x_C, x_A|\info(F)),
\end{align*}
where the last equality holds by Rule 3 in Theorem \ref{thm:gg_3_rules}, since
$B\indep_d N_{AC}|A, C$ in $\bar{G}$. Moreover, by (\ref{eq:sigma_generalized}) we have
\begin{align*}
P(x_C, x_A|\info(F))&=\sum_{x_B}\sum_{x_D}P(x_A|x_D)P(x_B|x_C, x_D)P(x_C|e_{AC})P(x_D)\\
&=P(x_C|e_{AC})\sum_{x_D}P(x_A|x_D)P(x_D)\sum_{x_B}P(x_B|x_C, x_D)\\
&=P(x_C|e_{AC})P(x_A),
\end{align*}
where $e_{AC}=g_{AC}(x_A)$. Hence, it follows that
\begin{align*}
P(x_B|\info(F)) =
\sum_{x_C}\sum_{x_A} P(x_C|e_{AC})P(x_B|x_C, x_A) P(x_A).
\end{align*}
\end{example}

\section{Concluding remarks}
This paper proposed a new \emph{info} intervention framework to formulate causal questions and
implement causal inferences in graphic models, including SCM and DAG.
In the \emph{info} intervention framework, the causality is viewed as information transfer, meaning that
$X$ causes $Y$ iff changing information on $X$ leads to a potential change in $Y$, while keeping everything else constant.
This new viewpoint allows us to do intervention by changing the information on $X$
to its descendants, making the counterfactual descendant variables of $X$ transfer this information, while
keeping the causal mechanisms of the model unchanged. Consequently, our information transfer causality
shares the features with interventionist causality, ``Law-like'' causality as well as
experimental causality.

In terms of communication, the \emph{info} intervention framework makes
the causal questions on non-manipulable variables non-controversially, and allows us to check the conditions related to
counterfactual variables visibly. In terms of theoretical focus, the causal calculus
based on ``back-door''/``front-door'' criteria and three rules is exchangeable with that
in Pearl's \emph{do} intervention framework, but under even simpler checking conditions.
As an extension, the generalized \emph{info} intervention framework was
studies to tackle more complicated causal questions, and this extension seems hard in the \emph{do} intervention framework.
Therefore, it is hoped that our \emph{info} intervention framework, as a standard for studying the causality, could be beneficial
to formalize, process and understand causal relationships in practice.

%



\setcounter{equation}{0}
\appendix

\section*{Appendix: Proofs}

To facilitate our proofs, we give a technical lemma which is analogous to the consistency assumption in the potential outcome framework (VanderWeele, 2009).

\begin{lem}
	\label{Lem:consistent}
	For an info-causal DAG $G$, $A, B$ and $C$ are its arbitrary disjoint node sets. Then,

(i) $P(x|\info(x_A)) = P(x)$;

(ii) $P(x_{A}, x_{B}| \info(x_A)) = P(x_{A}, x_{B})$;

(iii) $P(x_B|x_A, \info(x_A)) = P(x_B|x_A)$;

(iv) $P(x_A, x_B | \sigma(x_A), \sigma(x_C)) = P(x_A, x_B |\sigma(x_C))$;

(v) $P(x_B|x_A, \sigma(x_A), \sigma(x_C)) = P(x_B|x_A, \info(x_C))$.
\end{lem}

\textsc{Proof of Lemma \ref{Lem:consistent}.}
	The result (i) holds by taking $\tilde{x}_{A}=x_{A}$ in (\ref{eq:sigma}).
	By (i) and the marginalization over $x_A \cup x_B$ and $x_A$, it follows that
	\begin{align*}
	P(x_A, {x}_B|\info({x}_A)) = P(x_A, {x}_B)\,\,\,\mbox{ and }\,\,\,
	P({x}_A|\info({x}_A)) = P({x}_A),
	\end{align*}
   which entail the results (ii)--(iii). By (i), we know that
  $P(x | \sigma(x_A), \sigma(x_C)) = P(x | \sigma(x_C))= P(x)$, which entails the result (iv) by
  the marginalization over  $x_A \cup x_B$.  Finally, the result (v) holds by (iv) and a similar argument as for (iii).
  This completes all of the proofs.\qed

\vspace{4mm}

\textsc{Proof of Theorem \ref{back_doorthm}.}
	By (\ref{eq:sigma}), we have
 	$$P(x_{B}, x_{A}, x_{C}|\sigma(\tilde{x}_A)) = P(x_{B}|\tilde{x}_{A}, x_{C}) P(x_{A}|x_{C}) P(x_{C}),$$
  which entails
 	\begin{align*}
 		P(x_{B}|\info(\tilde{x}_A)) &= \sum_{x_{C}} \sum_{x_{A}} P(x_{B}, x_{A}, x_{C}|\sigma(\tilde{x}_A)) \\
 		&= \sum_{x_C} \sum_{x_A} P(x_B|\tilde{x}_A, x_C) P(x_A|x_C) P(x_C) \\
 		&= \sum_{x_C} P(x_B|\tilde{x}_A, x_C) P(x_C) \sum_{x_A} P(x_A|x_C) \\
 		&= \sum_{x_C} P(x_B|\tilde{x}_A, x_C) P(x_C).
 	\end{align*}
 	This completes the proof. \qed

\vspace{4mm}

\textsc{Proof of Theorem \ref{front_doorthm}.}
By (\ref{eq:sigma}), we have
	\begin{equation*}
	P(x_B, x_C, x_A, x_D| \info(\tilde{x}_A) )= P(x_B|x_D, x_C) P(x_C|\tilde{x}_A) P(x_A|x_D) P(x_D).
	\end{equation*}
Moreover, it is easy to see that $C\indep_d D|A$ and $A\indep_d B|C, D$ in Fig\,\ref{fig:front}.
In $G$, since the $d$-separation implies the conditional independence (Geiger, Verma and Pearl, 1990),
we know that $X_C$ and $X_D$ are independent given $X_A$, and
$X_A$ and $X_B$ are independent given $X_C$ and $X_D$. Hence,
\begin{equation}
	\label{eq:front:1}
P(x_C|x_D, x_A)=P(x_C|x_A) \,\,\,\mbox{and}\,\,\,P(x_B|x_D,x_C)=P(x_B|x_D,x_C,x_A),
\end{equation}
where the first equality further implies
\begin{equation}
	\label{eq:front:2}
P(x_D|x_C,x_A)=P(x_D|x_A).
\end{equation}
Then, it follows that
	\begin{align*}
	&P(x_B|\info(\tilde{x}_A)) \\
    &= \sum_{x_C} \sum_{x_A} \sum_{x_D} P(x_B, x_C, x_A, x_D| \info(\tilde{x}_A) )\\
	&= \sum_{x_C} \sum_{x_A} \sum_{x_D} P(x_B|x_D, x_C) P(x_C|\tilde{x}_A) P(x_A|x_D) P(x_D) \\
	&= \sum_{x_C} P(x_C|\tilde{x}_A) \sum_{x_A} \sum_{x_D} P(x_B|x_D,x_C)  P(x_A|x_D) P(x_D) \\
	&= \sum_{x_C} P(x_C|\tilde{x}_A) \sum_{x_A} \sum_{x_D} P(x_B|x_D,x_C, x_A)  P(x_D|x_A) P(x_A) \,\,\,\,\mbox{by }(\ref{eq:front:1})\\
	&= \sum_{x_C} P(x_C|\tilde{x}_A) \sum_{x_A} P(x_A) \sum_{x_D} P(x_B|x_D,x_C, x_A)  P(x_D|x_A) \\
	&= \sum_{x_C} P(x_C|\tilde{x}_A) \sum_{x_A} P(x_A) \sum_{x_D} P(x_B|x_D,x_C, x_A)  P(x_D|x_C, x_A) \,\,\,\,\mbox{by }(\ref{eq:front:2})\\
	&= \sum_{x_C} P(x_C|\tilde{x}_A) \sum_{x_A} P(x_A)  P(x_B|x_C, x_A).
	\end{align*}
This completes the proof. 	\qed

\vspace{4mm}

To prove Theorem \ref{thm:rules}, we first prove Theorem \ref{thm:rules:simple}.

\vspace{4mm}

\textsc{Proof of Theorem \ref{thm:rules:simple}.} In $G^{\info(\tilde{x}_A)}$, since $P(x_A, x_B, x_C, x_D|\info(\tilde{x}_A))$ factorizes, the $d$-separation implies the conditional independence (Geiger, Verma and Pearl, 1990).
	
	For Rule 1, we know that $X_B$ and $X_C$ are independent given $X_D$ in $G^{\info(\tilde{x}_A)}$, and hence the conclusion holds.
	
	For Rule 2, since $X_B$ and $X_A$ are independent given $X_C$ in $G^{\info(\tilde{x}_A)}$, we have that
$P(x_B|\info(\tilde{x}_A),x_C) = P(x_B|\info(\tilde{x}_A),\tilde{x}_A,x_C)$. Then, the conclusion holds since
	\begin{align*}
		P(x_B|\info(\tilde{x}_A),\tilde{x}_A,x_C)
		= \frac{P(x_B, \tilde{x}_A, x_C |\info(\tilde{x}_A))}{P(\tilde{x}_A, x_C |\info(\tilde{x}_A))}
		= \frac{P(x_B, \tilde{x}_A, x_C )}{P(\tilde{x}_A, x_C)},
	\end{align*}
where the second equality holds by Lemma \ref{Lem:consistent}(ii).

	For Rule 3, 
	let $Anc(B)=anc(B)\cup B$. Then, by (\ref{eq:sigma}), we have
	 \begin{align*}
	 	P(x| \sigma(\tilde{x}_A)) &= \prod_{k \in V} P(x_k|x_{pa(k)}^*) \\
	 	&= \prod_{k \in Anc(B)} P(x_k|x_{pa(k)}^*) \cdot  \prod_{k \notin Anc(B)} P(x_k|x_{pa(k)}^*) \\
	 	&= \prod_{k \in Anc(B)} P(x_k|x_{pa(k)}) \cdot  \prod_{k \notin Anc(B)} P(x_k|x_{pa(k)}^*),
	  \end{align*}
	  where we have used the fact that
	$x^*_{pa(k)} = x_{pa(k)}$ for any $k \in Anc(B)$, since $A \cap Anc(B)=\emptyset$.
	 Marginalizing over $x_{Anc(B)}$, we can obtain
	\begin{align*}
		P(x_{Anc(B)}| \sigma(\tilde{x}_A)) = \prod_{k \in Anc(B)} P(x_k|x_{pa(k)})
		= P(x_{Anc(B)}).
	\end{align*}	 	
Since $B \in Anc(B)$, the conclusion follows directly. This completes all of the proofs. \qed

\vspace{4mm}

For Theorem \ref{thm:rules}, its Rule 1 has been proved in Theorem \ref{thm:rules:simple}, and
its Rules 2 and 3 are proved below.

\vspace{4mm}

\textsc{Proof of Theorem \ref{thm:rules}} (Rule 2). Since $X_B$ and $X_C$ are independent given $X_D$ in $G^{\info(\tilde{x}_A, \tilde{x}_C)}$, we have that
$P(x_B|\info(\tilde{x}_A, \tilde{x}_C), x_D) = P(x_B|\info(\tilde{x}_A, \tilde{x}_C),$ $\tilde{x}_C, x_D)$. Then, the conclusion holds since
	\begin{align*}
		P(x_B|\info(\tilde{x}_A, \tilde{x}_C), \tilde{x}_C, x_D)
		&= \frac{P(x_B, x_D |\info(\tilde{x}_A, \tilde{x}_C), \tilde{x}_C)}{P(x_D |\info(\tilde{x}_A, \tilde{x}_C), \tilde{x}_C)}\\
		&= \frac{P(x_B, x_D |\info(\tilde{x}_A), \tilde{x}_C)}{P(x_D |\info(\tilde{x}_A), \tilde{x}_C)},
	\end{align*}
where the last equality holds by Lemma \ref{Lem:consistent}(v). \qed

\vspace{4mm}

To prove Rule 3 in Theorem \ref{thm:rules}, we need an additional lemma.

\begin{lem}
For an info-causal DAG $G$, $B, C_1$, $C_2$ and $D$ are its arbitrary disjoint node sets. Then,

(i) $P(x_{B}|\info(\tilde{x}_{C_1}), \info(\tilde{x}_{C_2}), x_{D}) = P(x_{B}| \info(\tilde{x}_{C_2}), x_{D})$ if $B \indep_d C_1|D$ in $G^{\info(\tilde{x}_{C_2})}$;

(ii) $P(x_{B}|\info(\tilde{x}_{C_2}), x_D) = P(x_B| x_D)$ if there are no causal paths from $C_2$ to $B\cup D$ in $G$. \label{lem_A2}
\end{lem}

\textsc{Proof of Lemma \ref{lem_A2}.} First, since $B \indep_d C_1|D$ in $G^{\info(\tilde{x}_{C_2})}$,
 we know that $B \indep_d C_1|D$ in $G^{\info(\tilde{x}_{C_1}, \tilde{x}_{C_2})}$.
 Then, the result (i) follows by the fact that
 \begin{align*}
& P(x_{B}|\info(\tilde{x}_{C_1}), \info(\tilde{x}_{C_2}), x_{D}) \\
 &= P(x_{B}|\tilde{x}_{C_1}, \info(\tilde{x}_{C_2}), x_{D}) \,\,\,\mbox{ (by Rule 2 in Theorem } \ref{thm:rules})\\
 & = P(x_{B}|\info(\tilde{x}_{C_2}), x_{D}) \,\,\,\mbox{ (by Rule 1 in Theorem } \ref{thm:rules}).
 \end{align*}
Second, since there are no causal paths from $C_2$ to $B\cup D$, by Rule 3 in Theorem  \ref{thm:rules:simple} we have
\begin{align*}
P(x_{B}, x_{D}|\info(\tilde{x}_{C_2})) = P(x_B, x_D)\,\,\mbox{ and }\,\,
P(x_{D}|\info(\tilde{x}_{C_2})) = P(x_D),
\end{align*}
which entail that the result (ii) holds. This completes all of the proofs. \qed

\vspace{4mm}

\textsc{Proof of Theorem \ref{thm:rules}} (Rule 3). Let $C_1 = C\cap anc(D)$ and $C_2 = C/anc(D)$. It suffices to show
\begin{align}
P(x_{B}|\info(\tilde{x}_{C_1}), \info(\tilde{x}_{C_2}), x_{D}) = P(x_{B}|w_D),\,\,\ \mbox{ if } B \indep_d C|D \mbox{ in }G_{\overline{C_2}}.
\end{align}

First, we prove that if $B \indep_d C|D$ in $G_{\overline{C_2}}$, then
\begin{align}\label{aaa_1}
B \indep_d C_1|D \mbox{ in } G^{\info(\tilde{x}_{C_2})},
\end{align}
and hence by Lemma \ref{lem_A2}(i) we have
\begin{align}\label{aaa_2}
P(x_{B}|\info(\tilde{x}_{C_1}), \info(\tilde{x}_{C_2}), x_{D})=P(x_{B}| \info(\tilde{x}_{C_2}), x_{D}).
\end{align}
Suppose the result (\ref{aaa_1}) does not hold. Then, there exists a $D$-connected path from $B$ to $C_1$ in $G^{\info(\tilde{x}_{C_2})}$. Note that this path can not contain any node in $C_2$.
This is because if this path includes a node $c^{*}\in C_2$,
then $c^{*}\not\in anc(D)$ must be a collider, in view of the fact that the nodes $C_2$ in $G^{\info(\tilde{x}_{C_2})}$ have no output edges.
It turns out that this path is blocked by $D$, leading to a contradiction.
Therefore, since this $D$-connected path does not contain any node in $C_2$, it is also in $G_{\overline{C_2}}$, leading to a contradiction with the condition that
$B \indep_d C|D$ in $G_{\overline{C_2}}$

Second, we prove that if $B \indep_d C|D$ in $G_{\overline{C_2}}$, then
\begin{align}\label{aaa_3}
\mbox{there are no causal paths from } C_2 \mbox{ to }B \mbox{ in }G,
\end{align}
and hence by Lemma \ref{lem_A2}(ii) and the fact that $C_2\cap anc(D)=\emptyset$, we have
\begin{align}\label{aaa_4}
P(x_{B}| \info(\tilde{x}_{C_2}), x_{D})=P(x_B| x_D).
\end{align}
Suppose the result (\ref{aaa_3}) does not hold. Then, there exists a shortest causal path from $C_2$ to $B$ in $G$,
and this shortest path contains only one node in $C_2$. Hence, this shortest path is also in
$G_{\overline{C_2}}$. Since $B \indep_d C|D$ in $G_{\overline{C_2}}$, it implies that $C_2\cap anc(D)\not= \emptyset$,
leading to a contradiction with the fact that $C_2\cap anc(D)=\emptyset$.

Finally, the conclusion follows by (\ref{aaa_2}) and (\ref{aaa_4}). \qed

\vspace{4mm}

\textsc{Proof of Theorem \ref{Equivalence}.}  We first prove that if
 $B \indep_d C|A, D$ in $G_{\overline{A}}$, then
 \begin{align}\label{aaa_5}
 B \indep_d C|D \mbox{ in } G^{\info(\tilde{x}_{A})}.
 \end{align}
To prove (\ref{aaa_5}), it suffices to show that any path $\ell$ from $B$ to $C$ in $G^{\info(\tilde{x}_{A})}$ is blocked by $D$.
We consider two different cases:

Case I: if the path $\ell$ contains a node $a^*\in A$, then $a^*\not\in anc(D)$ must be a collider in $G^{\info(\tilde{x}_{A})}$, since the nodes $A$ in $G^{\info(\tilde{x}_{A})}$
have no output edges. Hence, the path $\ell$ is blocked by $D$ in Case I.

Case II: if the path $\ell$ contains no nodes in $A$, then $\ell$ is also a path in $G_{\overline{A}}$, and hence
it is blocked by $A$ and $D$ in $G_{\overline{A}}$, due to the condition that $B \indep_d C|A, D$ in $G_{\overline{A}}$.
In other words, there exists a node $\kappa$, which blocks this path $\ell$ in $G_{\overline{A}}$. If
$\kappa$ is a collider, then $\kappa\not\in anc(A\cup D)$ in $G_{\overline{A}}$, indicating that there has
no causal path from $\kappa$ to $D$ in $G_{\overline{A}}$. Then, it further implies that
there has no causal path from $\kappa$ to $D$ in $G^{\info(\tilde{x}_{A})}$, meaning that
the path $\ell$ is blocked by $D$ in $G^{\info(\tilde{x}_{A})}$.

If $\kappa$ is not a collider, then $\kappa\in A\cup D$ in $G_{\overline{A}}$. Since the path $\ell$ contains no nodes in $A$, it follows that
$\kappa\in D$ in $G^{\info(\tilde{x}_{A})}$, meaning that
the path $\ell$ is blocked by $D$ in $G^{\info(\tilde{x}_{A})}$.

Overall, we have shown that no matter whether $\kappa$ is a collider, the path $\ell$ is blocked by $D$ in Case II.
Therefore, the result (\ref{aaa_5}) holds. Similarly, we can show that
if $B \indep_d C|D$ in $G^{\info(\tilde{x}_{A})}$, then
 $B \indep_d C|A, D$ in $G_{\overline{A}}$. Hence, the result (i) holds.

Note that the nodes $C$ are chosen arbitrarily in the proof of (i). So,  the results (ii)--(iii) follow by
 the same argument as for the result (i). This completes all of the proofs. \qed

\end{document}